%% file: main.tex
 \documentclass[pmlr,twocolumn,10pt]{jmlr} 

\usepackage{booktabs}
\usepackage{siunitx}

\usepackage{arydshln}
\usepackage{multirow,xspace,pifont}
\newcommand{\methodname}{U-Fair\xspace}
\newcommand{\xmark}{\ding{53}}%

\usepackage{graphicx}
\usepackage{comment}

\newcommand{\Loss}{\mathcal{L}}
\newcommand{\x}{\mathbf{x}}
\usepackage{soul}

\usepackage[switch]{lineno}

\newcommand{\equal}[1]{{\hypersetup{linkcolor=black}\thanks{#1}}}

\theorembodyfont{\upshape}
\theoremheaderfont{\scshape}
\theorempostheader{:}
\theoremsep{\newline}

\jmlrvolume{259}
\jmlryear{2024}
\jmlrsubmitted{LEAVE UNSET}
\jmlrpublished{LEAVE UNSET}
\jmlrworkshop{Machine Learning for Health (ML4H) 2024} 

\title[U-Fair: Uncertainty-based Multimodal Multitask Learning for Fairer Depression Detection]{U-Fair: Uncertainty-based Multimodal Multitask Learning \\ for Fairer Depression Detection}

\author{
\Name{Jiaee Cheong}\equal{This work was undertaken while Jiaee Cheong was a visiting PhD student at METU.} 
\Email{jc2208@cam.ac.uk}\\
\addr University of Cambridge \& the Alan Turing Institute, United Kingdom.
\AND
\Name{Aditya Bangar} 
\Email{adityavb21@iitk.ac.in}\\
\addr Indian Institute of Technology, Kanpur, India.
\AND
\Name{Sinan Kalkan}
\Email{skalkan@metu.edu.tr}\\
\addr Dept. of Comp. Engineering and ROMER Center for Robotics and AI, \\
Middle East Technical University (METU), Turkiye.
\AND
\Name{Hatice Gunes} 
\Email{hg410@cam.ac.uk}\\
\addr University of Cambridge, United Kingdom.
}
\begin{document}

\maketitle

\begin{abstract}
\input{Main_Content/0_Abstract}
\end{abstract}

\input{Main_Content/1_Introduction}

\input{Main_Content/2_Literature_Review}

\input{Main_Content/3_Methods}

\input{Main_Content/4_Experiments}

\input{Main_Content/5_Results}

\input{Main_Content/6_Discussion}

\clearpage

\acks{
\noindent\textbf{Funding:} 
J. Cheong is supported by the Alan Turing Institute doctoral studentship, the Leverhulme Trust and further acknowledges resource support from METU. 
A. Bangar contributed to this while undertaking a remote visiting studentship at the Department of Computer Science and Technology, University of Cambridge.
H. Gunes’ work is supported by the EPSRC/UKRI project ARoEq under grant ref. EP/R030782/1.
\noindent\textbf{Open access:} 
The authors have applied a Creative Commons Attribution (CC BY) licence to any Author Accepted Manuscript version arising.
\noindent\textbf{Data access:} 
This study involved secondary analyses of existing datasets. All datasets are described and cited accordingly. 
}

\bibliography{references}

\clearpage

\appendix

\input{Main_Content/appendix}

\end{document}

%% file: Main_Content/0_Abstract.tex

Machine learning bias in mental health is becoming an increasingly pertinent challenge. 
Despite promising efforts indicating that multitask approaches often work better than unitask approaches, there is minimal work investigating the impact of multitask learning on performance and fairness in depression detection nor leveraged it to achieve fairer prediction outcomes.
In this work, we undertake a systematic investigation of using a multitask approach to improve performance and fairness for depression detection. 
We propose a novel gender-based task-reweighting method using uncertainty grounded in how the PHQ-8 questionnaire is structured.
Our results indicate that, 
although a multitask approach improves performance and fairness compared to a unitask approach, the results are not always consistent and we see evidence of negative transfer and a reduction in the Pareto frontier, which is concerning given the high-stake healthcare setting.
Our proposed approach of gender-based reweighting with uncertainty improves performance and fairness and alleviates both challenges to a certain extent.
Our findings on each PHQ-8 subitem \textit{task difficulty} are also in agreement with the largest study conducted on the PHQ-8 subitem \textit{discrimination capacity}, thus providing the very first tangible evidence linking ML findings with large-scale empirical population studies conducted on the PHQ-8. 

%% file: Main_Content/1_Introduction.tex
%
\section{Introduction}

\begin{figure*}
    \centering
    \includegraphics[width=1.0\textwidth]{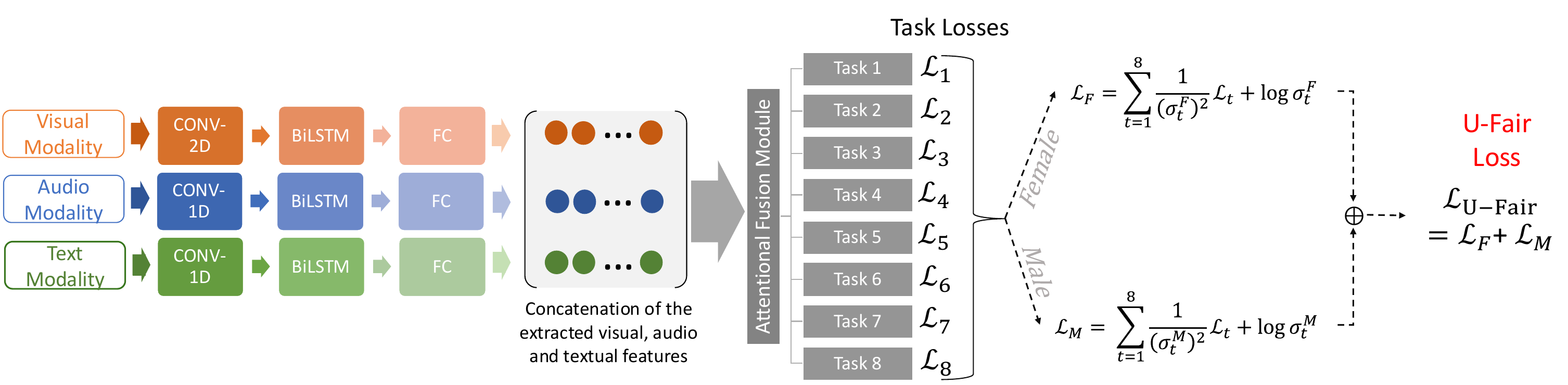}
    \vspace{-0.7cm}
    \caption{Our proposed method is rooted in the observation that each gender may have different PHQ-8 distributions and different levels of task difficulty across the $t_1$ to $t_8$ tasks. We propose accounting for this gender difference in PHQ-8 distributions via \methodname.}
    \label{fig:MTL_Overview}
   \vspace{-0.5cm}
\end{figure*}

Mental health disorders (MHDs) are becoming increasingly prevalent world-wide 
\citep{wang2007use} 
Machine learning (ML) methods have been successfully applied to many real-world and health-related  areas \citep{sendak2020human}. The natural extension of using ML for MHD analysis and detection has proven to be promising \citep{long2022scoping,he2022deep,zhang2020multimodal}.
On the other hand, ML bias is becoming an increasing source of concern \citep{buolamwini2018gender,barocas2017fairness,xu2020investigating,cheong2021hitchhiker,cheong2022counterfactual,cheong2023causal}. Given the high stakes involved in MHD analysis and prediction, it is crucial to investigate and mitigate the ML biases present. 
A substantial amount of literature has indicated that adopting a multitask learning (MTL) approach towards depression detection demonstrated significant improvement across classification-based performances \citep{
li2022multi,zhang2020multimodal}.
Most of the existing work rely on the standardised 
and commonly used eight-item Patient Health Questionnaire depression scale (PHQ-8) \citep{kroenke2009phq} to obtain the ground-truth labels on whether a subject is considered depressed. 
A crucial observation is that in order to arrive at the final classification (depressed vs non-depressed), 
a clinician has to first obtain the scores of each of the PHQ-8 sub-criterion and then sum them up to arrive at the final binary classification (depressed vs non-depressed).
Details on how the final score is derived from the PHQ-8 questionnaire can be found in Section \ref{sect:problem_form}.

Moreover, each gender may display different PHQ-8 task distribution which may results in different PHQ-8 distribution and variance. 
Although investigation on the relationship between the PHQ-8 and gender has been explored in other fields such as psychiatry 
\citep{
thibodeau2014phq,vetter2013gender,leung2020measurement},
this has not been investigated nor accounted for in any of the existing ML for depression detection methods. 
Moreover, 
existing work has demonstrated the risk of a fairness-accuracy trade-off \citep{pleiss2017fairness} and how mainstream MTL objectives might not correlate well with fairness goals \citep{wang2021understanding}. 
No work has investigated how a MTL approach impacts performance across fairness for the task of \textit{depression detection}.

In addition, prior works have demonstrated the intricate relationship between ML bias and uncertainty \citep{mehta2023evaluating,tahir2023fairness,kaiser2022uncertainty,kuzucu2024uncertainty}.
Uncertainty broadly refers to confidence in predictions.
Within ML research, two types of uncertainty are commonly studied: data (or
aleatoric) and model (or epistemic) uncertainties. 
Aleatoric uncertainty refers to the inherent randomness 
in the experimental outcome whereas
epistemic uncertainty can be attributed to a lack of knowledge \citep{gal_uncertainty}. 
A particularly relevant theme 
is that ML bias can be attributed to uncertainty in some models or datasets \citep{kuzucu2024uncertainty} 
and that taking into account uncertainty as a bias mitigation strategy has proven effective \citep{tahir2023fairness,kaiser2022uncertainty}.
A growing body of literature has also highlighted the importance of taking uncertainty into account within
a range of tasks \citep{naik2024pre,han2024perspectives,baltaci2023class,cetinkaya2024ranked} and
healthcare settings \citep{grote2022enabling,chua2023tackling}.
Motivated by the above and 
the importance of a clinician-centred approach towards building relevant ML for healthcare solutions, 
we propose a novel method, \methodname, which accounts for the gender difference in PHQ-8 distribution and
leverages on uncertainty as
a MTL task reweighing mechanism to achieve better gender fairness for depression detection. 
Our key contributions are 
as follow:
\begin{itemize}
     \vspace{-2mm}\item We conduct the first analysis to investigate how MTL impacts fairness in \textit{depression detection} by using each PHQ-8 subcriterion as a task. We show that a simplistic baseline MTL approach runs the risk of incurring negative transfer and may not improve on the Pareto frontier. 
    A Pareto frontier can be understood as the set of optimal solutions that strike a balance among different objectives such that there is no better solution beyond the frontier.
   \vspace{-2mm}\item We propose a simple yet effective approach that leverages gender-based aleatoric uncertainty which improves the fairness-accuracy trade-off and alleviates the negative transfer phenomena and improves on the Pareto-frontier beyond a unitask method. 
    \vspace{-2mm}\item We provide the very first results connecting 
    the empirical results obtained via ML experiments with the \textit{empirical findings} obtained via the \textit{largest study conducted on the PHQ-8}. Interestingly, our results highlight the intrinsic relationship between task difficulty as quantified by aleatoric uncertainty and the discrimination capacity of each item of the PHQ-8 subcriterion. 

\end{itemize}

\begin{table*}[ht!]
  \centering
  \footnotesize
  \addtolength{\tabcolsep}{-0.7mm}
  \begin{tabular}{l|c|c|cc|ccc}
    \hline
     & &  
     & \multicolumn{2}{c|}{\textbf{Approach}} 
     &\multicolumn{3}{c}{\textbf{Evaluation}} \\

    {\textbf{Study}} & \textbf{Problem} & \textbf{Multimodal}     &Uncertainty  &NFM Measures  &PF &NT  &ND  \\ 
    \hline

     \cite{zanna2022bias}   &Anxiety        & \xmark &\checkmark  &2  &\xmark & \xmark  &1   \\ 

     \cite{li2023multi} &Healthcare prediction  & \xmark & \xmark &2 &\xmark & \xmark &1\\ 
     \cite{li2023feri} &Organ transplant        & \xmark & \xmark  &2 &\xmark & \xmark &1\\
     \cite{ban2024fair} &Resource allocation    & \xmark & \xmark  &2 &\checkmark & \xmark &3\\
     \cite{li2024transformer} &Risk factor prediction  & \xmark & \xmark & 2  & \xmark & \xmark &1\\
    \cdashline{1-8}[.4pt/2pt]%
     
     \methodname (\textbf{Ours}) &Depression detection & \checkmark (AVT) &\checkmark  &4 &\checkmark & \xmark &2 \\

   \hline

    \hline
  \end{tabular}

  \caption{Comparative Summary with existing MTL Fairness studies. Abbreviations (sorted): A: Audio. NFM: Number of Fairness Measures. NT: Negative Transfers. ND: Number of Datasets. PF: Pareto Frontier. T: Text. V: Visual. 
  }   \label{tab:comparative_summary}
  \vspace{-0.5cm}
\end{table*}
%

%% file: Main_Content/2_Literature_Review.tex

\section{Literature Review}

Gender difference in depression manifestation has long been studied and recognised within fields such as medicine \citep{barsky2001somatic} and psychology \citep{hall2022systematic}. 
Anecdotal evidence has also often supported this view. Literature indicates that females and males tend to show different behavioural symptoms when depressed \citep{barsky2001somatic,ogrodniczuk2011men}. 
For instance, certain acoustic features (e.g. MFCC) are only statistically significantly different between depressed and healthy males \citep{wang2019acoustic}. On the other hand, compared to males, depressed females are more emotionally expressive and willing to reveal distress via behavioural cues \citep{barsky2001somatic,jansz2000masculine}.

Recent works have indicated that ML bias is present within 
mental health analysis
\citep{zanna2022bias,bailey2021gender,cheong_fairrefuse,cheong2024_tac,cameron2024multimodal,spitale2024underneath}.
\citet{zanna2022bias} proposed an uncertainty-based approach to address the bias present in the TILES dataset.
\citet{bailey2021gender} demonstrated the effectiveness of using an existing bias mitigation method, data re-distribution, to mitigate the gender bias present in the DAIC-WOZ dataset.
\citet{cheong_gender_fairness,cheong_fairrefuse} demonstrated that bias exists in existing mental health algorithms and datasets and subsequently proposed a causal multimodal method to mitigate the bias present.

MTL is noted to be particularly effective when the tasks are correlated \citep{zhang2021_multitask_survey}.
Existing works using MTL for depression detection has proven fruitful.
\citet{ghosh2022multitask} adopted a MTL approach by training the network to detect three closely related tasks: depression, sentiment and emotion.
\citet{wang2022online} proposed a MTL approach using word vectors and statistical features.
\citet{li2022multi} 
implemented a similar strategy by using depression and three other auxiliary tasks: topic, emotion and dialog act.
\citet{gupta2023multimodal} adopted a multimodal, multiview and MTL approach where the subtasks are depression, sentiment and emotion.

In concurrence, although MTL has proven to be effective at improving \textit{fairness} for other tasks such as 
healthcare predictive modelling \citep{li2023multi}, organ transplantation \citep{li2023feri} and resource allocation \citep{ban2024fair}, 
this approach has been underexplored for the task of depression detection.

\paragraph{Comparative Summary:} Our work differs from the above in the following ways (see Table \ref{tab:comparative_summary}). First, our work is the first to leverage an MTL approach to improve gender fairness in \textit{depression detection}.
Second, we utilise an MTL approach where each task corresponds to each of the PHQ-8 subtasks \citep{kroenke2009phq} in order to exploit gender-specific differences in PHQ-8 distribution to achieve greater fairness.
Third, we propose a novel gender-based uncertainty MTL loss reweighing to achieve fairer performance across gender for

%% file: Main_Content/3_Methods.tex
\section{Methodology: \methodname}

In this section, we introduce \methodname, which 
uses aleatoric-uncertainties for demographic groups to reweight their losses.  

\subsection{PHQ-8 Details}
\label{sect:problem_form}

One of the standardised and most commonly used depression evaluation method is the PHQ-8 developed by \citet{kroenke2009phq}.
In order to arrive at the final classification (depressed vs non-depressed), the protocol is to first obtain the subscores of each of the PHQ-8 subitem
as follows:
\begin{itemize}
\vspace{-2mm}
\item PHQ-1: Little interest or pleasure in doing things, \vspace{-2mm}
\item PHQ-2: Feeling down, depressed, or hopeless, \vspace{-2mm}
\item PHQ-3: Trouble falling or staying asleep, or sleeping too much, \vspace{-2mm}
\item PHQ-4: Feeling tired or having little energy, \vspace{-2mm}
\item PHQ-5: Poor appetite or overeating, \vspace{-2mm}
\item PHQ-6: Feeling 
that you are a failure, \vspace{-2mm} 
\item PHQ-7: Trouble concentrating on things, \vspace{-2mm}
\item PHQ-8: Moving or speaking so slowly that other people could have noticed. 
\end{itemize}

Each PHQ-8 subcategory is scored between $0$ to $3$, with the final PHQ-8 total score (TS) ranging between $0$ to $24$. 
The PHQ-8 binary outcome is obtained via thresholding.
A PHQ-8 TS of $\geq 10$ belongs to the depressed class ($Y=1$) whereas TS $\leq10$ belongs to the non-depressed class ($Y=0$). 

Most existing 
works focused on predicting the final binary class ($Y$) \citep{zheng2023two,bailey2021gender}.
Some focused on predicting the PHQ-8 total score and further obtained the binary classification via thresholding according to the formal definition \citep{williamson2016detecting,gong2017topic}. Others adopted a bimodal setup with 2 different output heads to predict the PHQ-8 total score as well as the PHQ-8 binary outcome \citep{valstar2016avec,al2018detecting}.

\subsection{Problem Formulation}

In our work, \textit{in alignment with how the PHQ-8 works}, we adopt the approach where each 
PHQ-8 subcategory is treated as a task $t$. 
The architecture is adapted from \citet{wei2022multi}. 
For each individual $i \in I$,
we have 8 different prediction heads for each of the tasks, [$t_1$, ..., $t_8$] $\in T$, to predict the score $y_t^i\in \{0,1,2,3\}$  for each task or sub PHQ-8 category. 
The ground-truth labels for each task $t$ is transformed into a Gaussian-based soft-distribution $p_t (x)$,
as soft labels provide more information for the model to learn from \citep{yuan2024learning}. 
$x$ is the input feature provided to the model.
Each of the classification heads are trained to predict the probability $q_t (x)$ of the 4 different score classes $y_t^i\in \{0,1,2,3\}$. During inference, the final $y_t^i\in \{0,1,2,3\}$ is obtained by selecting the score with the maximum probability.
The PHQ-8 Total Score $TS$ and final PHQ-8 binary classification $\hat{Y}$ for each individual $i \in I $ are derived from each subtask via: 
\begin{equation}
TS = \sum_{t=1}^{8} y_t,
\end{equation}
and
\begin{equation}
\hat{Y} = 1       \text{ if } TS \geq 10,  \text{ else } \hat{Y}=0.
\end{equation}
$\hat{Y}$ thus denotes the final predicted class calculated based on the summation of $y_t$.
We study the problem of fairness in depression detection, where the goal is to predict a correct outcome $y^i\in Y$ from input $\mathbf{x}^i\in X$ based on the available dataset $D$ for individual $i\in I$.
In our setup, $Y=1$ denotes the PHQ-8 binary outcome corresponding to ``depressed" and $Y=0$ denotes otherwise.
Only gender was provided as a sensitive attribute $S$.


\subsection{Unitask Approach}
\label{sect:unitask_approach}
For our single task approach, we use a Kullback-Leibler (KL) Divergence loss as follows: 
\begin{equation}
    \Loss_{STL}
    = 
    \sum_{t\in T} 
    p_t (x) \log \left( \frac{p_t (x)}{q_t (x)} \right) . 
    \label{eqn:STL_loss}
\end{equation}
$p_t (x)$ is the soft ground-truth label for each task $t$ and $q_t (x)$ is the probability of the $4$ different score classes $y_t\in \{0,1,2,3\}$ as explained in Section \ref{sect:problem_form}.

\subsection{Multitask Approach}
For our baseline multitask approach, we extend the loss function in Equation \ref{eqn:STL_loss} to arrive at the following generalisation:
\begin{equation}
    \Loss_{MTL} = \sum_{t\in T} w_t \Loss_t.
    \label{eqn:MTL_loss}
\end{equation}
$\Loss_t$ is the single task loss $\Loss_{STL}$ for each $t$ as defined in Equation \ref{eqn:STL_loss}.
We set $w_t=1$ in our experiments.

\subsection{Baseline Approach}
To compare between the generic multitask approach in Equation \ref{eqn:MTL_loss} and an \textit{uncertainty-based} loss reweighting approach, 
we use the commonly used multitask learning method by \citet{kendall2018multi} as the baseline uncertainty weighting (UW) appraoch. The uncertainty MTL loss across tasks is thus defined by: 
\begin{equation}
    \Loss_{UW} = \sum_{t\in T} \left(\frac{1}{\sigma_t^2} \Loss_t + \log\sigma_t \right), 
    \label{eqn:UW_MTL}
\end{equation}
where $\Loss_t$ is the single task loss as defined in Equation \ref{eqn:STL_loss}. $\sigma_t$ is the learned weight of loss for each task $t$ and can be interpreted as the aleatoric uncertainty of the task. 
A task with a higher aleatoric uncertainty will thus lead to a larger single task loss $\Loss_t$ thus preventing the trained model to optimise on that task. 
The higher $\sigma_t$, the more difficult the task $t$. 
$\log\sigma_t$ penalizes the model from arbitrarily increasing $\sigma_t$ to reduce the overall loss \citep{kendall2018multi}.


\subsection{Proposed Loss: \methodname}
To achieve fairness across the different PHQ-8 tasks, we propose the idea of task prioritisation based on the model's task-specific uncertainty weightings.
%
Motivated by literature highlighting the existence of gender difference in depression manifestation \citep{barsky2001somatic},
we propose a novel gender based uncertainty reweighting approach and introduce \methodname Loss which is defined as follows:
\begin{equation}
    \Loss_{\methodname} =\frac{1}{|S|} \sum_{s\in S} \sum_{t\in T} \left( \frac{1}{\left(\sigma_t^s \right)^2} \Loss_t^s + \log \sigma_t^s \right).
\end{equation}
For our setting, $s$ can either be male $s_1$ or female $s_0$ and $|S|=2$. 
Thus, we have the uncertainty weighted task loss for each gender, and sum them up to arrive at our proposed loss function $\Loss_{MMFair}$.

This methodology has two key benefits. 
First, fairness is optimised implicitly as we train the model to optimise for task-wise prediction accuracy. %
As a result, by not constraining the loss function to blindly optimise for fairness at the cost of utility or accuracy, we hope to reduce the negative impact on fairness and improve the Pareto frontier with a constraint-based fairness optimisation approach \citep{wang2021understanding}. 
Second, as highlighted by literature in psychiatry \citep{leung2020measurement,thibodeau2014phq},
each task has different levels of uncertainty in relation to each gender. By adopting a 
gender based uncertainty loss-reweighting approach, we account for such uncertainty in a principled manner, thus encouraging the network to learn a better \textit{joint-representation} due to the MTL and the gender-base aleatoric uncertainty loss reweighing approach.

%% file: Main_Content/4_Experiments.tex
\section{Experimental Setup}

We outline the implementation details and evaluation measures here. 
We use DAIC-WOZ \citep{valstar2016avec} and E-DAIC 
\citep{ringeval2019avec} for our experiments.
Further details about the datasets can be found within the Appendix. 

\subsection{Implementation Details}
We adopt an attention-based multimodal architecture adapted from \citet{wei2022multi}
featuring late fusion of extracted representations from the three different modalities (audio, visual, textual) as illustrated in Figure \ref{fig:MTL_Overview}.
The extracted features from each modality are concatenated in parallel to form a feature map as input to the subsequent fusion layer. We have 8 different attention fusion layers connected to the 8 output heads which corresponds to the $t_1$ to $t_8$ tasks.
For all loss functions, we train the models with the Adam optimizer  \citep{kingma2014adam} at a learning rate of 0.0002 and a batch size of 32. 
We train the network for a maximum of 150 epochs and apply early stopping.

\subsection{Evaluation Measures} 
To evaluate performance, we use F1, recall, precision, accuracy and unweighted average recall (UAR)
in accordance with existing work \citep{cheong_acii_23}.
%
To evaluate group fairness, we use the most commonly-used definitions according to \citep{hort2022bias}. 
$s_1$ denotes the male majority group and $s_0$ denotes the female minority group for both datasets.
%
%
%

%
\begin{itemize}   
\item\textbf{Statistical Parity}, or demographic parity, is based purely on predicted outcome $\hat{Y}$ and independent of actual outcome $Y$:
    \begin{equation}
    \label{eqn:SP}
     \mathcal{M}_{SP}= \frac{P(\hat{Y}=1|s_0 ) }{ P(\hat{Y}=1 | s_1)} .
    \end{equation}
    According to $\mathcal{M}_{SP}$, in order for a classifier to be deemed fair, $P(\hat{Y}=1 | s_1) = P(\hat{Y}=1|s_0 ) $.

\item\textbf{Equal opportunity} states that both demographic groups $s_0$ and $s_1$ should have equal True Positive Rate (TPR). 
    \begin{equation}
    \label{eqn:EOpp}
    \mathcal{M}_{EOpp} = \frac{P(\hat{Y}=1|Y=1, s_0 )}{P(\hat{Y}=1 | Y=1, s_1)}.
    \end{equation}
    According to this measure, in order for a classifier to be deemed fair,  $P(\hat{Y}=1 | Y=1, s_1) = P(\hat{Y}=1|Y=1, s_0 ) $.
     
\item\textbf{Equalised odds} can be considered as a generalization of Equal Opportunity where the rates are not only equal for $Y=1$, but for all values of $Y \in \{1, ... k\}$, i.e.: 
    \begin{equation}
        \label{eqn:EOdd}
        \mathcal{M}_{EOdd} = \frac{P(\hat{Y}=1|Y=i, s_0 )}{P(\hat{Y}=1 | Y=i, s_1)} .
    \end{equation}
    According to this measure, in order for a classifier to be deemed fair, $P(\hat{Y}=1 | Y=i, s_1) = P(\hat{Y}=1|Y=i, s_0 ),  \forall  i \in \{1, ... k\}$.

\item\textbf{Equal Accuracy} states that both subgroups $s_0$ and $s_1$ should have equal rates of accuracy.
    \begin{equation}
        \label{eqn:Wacc}
        \mathcal{M}_{EAcc} = \frac{\mathcal{M}_{ACC,s_0}}{\mathcal{M}_{ACC,s_1}} .
    \end{equation}
\end{itemize}
For all fairness measures, 
the ideal score of $1$ thus indicates that both measures are equal for $s_0$ and $s_1$ and is thus considered ``perfectly fair''. 
We adopt the approach of existing work which considers $0.80$ and $1.20$ as the 
lower and upper fairness bounds respectively \citep{zanna2022bias}.
Values closer to $1$ are fairer, values further form $1$ are less fair.
%
For all binary classification, the ``default" threshold of $0.5$ is used in alignment with existing works \citep{wei2022multi,zheng2023two}.

\begin{table}[ht!]
  \setlength{\tabcolsep}{2.5pt}
\small
  \centering
  \begin{tabular}{l|l|l|c}
    \hline
   \phantom{......} & \textbf{Measure}  &\textbf{Approach} & \textbf{Binary Outcome} \\

    \hline

  \parbox[t]{2mm}{\multirow{20}{*}{\rotatebox[origin=c]{90}{Performance Measures}}} &  \multirow{4}{*}{Acc}     &Unitask       &0.66    \\
    &        &Multitask      &0.70             \\
    &        &Baseline UW  	&\textbf{0.82}    \\
     \cdashline{3-4}[.4pt/2pt] &       
            & \methodname \textbf{(Ours)}      &0.80   \\

    \cline{2-4}
    & \multirow{4}{*}{F1}          &Unitask             &0.47 \\
    &            &Multitask           &0.53 \\
     &           &Baseline UW         	&0.29    \\
     \cdashline{3-4}[.4pt/2pt] &           
                &\methodname \textbf{(Ours)}            	 &\textbf{0.54}   \\
                
    \cline{2-4}
    & \multirow{4}{*}{Precision}   &Unitask      &0.44 \\
    &             &Multitask      &0.50 \\
     &           &Baseline UW   	&0.22 \\
     \cdashline{3-4}[.4pt/2pt] &           
                &\methodname \textbf{(Ours)}     &\textbf{0.56}\\
                
    \cline{2-4}
    & \multirow{4}{*}{Recall}      &Unitask      &0.50 \\
     &           &Multitask   &0.57 \\
     &           &Baseline UW   	&0.43  \\
     \cdashline{3-4}[.4pt/2pt] &           
                &\methodname \textbf{(Ours)}   &\textbf{0.60}  \\
                
    \cline{2-4}
    & \multirow{4}{*}{UAR}         &Unitask   &0.60 \\
     &            &Multitask  &\textbf{0.65} \\
     &           &Baseline UW  	&0.64 \\
     \cdashline{3-4}[.4pt/2pt] &           
                &\methodname \textbf{(Ours)} 	 &0.63\\
     \hline

    \parbox[t]{2mm}{\multirow{16}{*}{\rotatebox[origin=c]{90}{Fairness Measures}}} & \multirow{4}{*}{\textbf{\(\mathcal{M}_{SP}\)}} &Unitask    &0.47 \\
     &           &Multitask  &0.86 \\
     &           &Baseline UW  	&1.23\\
     \cdashline{3-4}[.4pt/2pt] &           
                &\methodname \textbf{(Ours)} 	&\textbf{1.06}\\
    \cline{2-4}            
    & \multirow{4}{*}{\textbf{\(\mathcal{M}_{EOpp}\)}}   &Unitask  &0.45 \\
     &           &Multitask  &\textbf{0.78} \\
     &           &Baseline UW  	&1.70 \\
     \cdashline{3-4}[.4pt/2pt] &           
                &\methodname \textbf{(Ours)}  	&1.46 \\
            
    \cline{2-4}       
    & \multirow{4}{*}{\textbf{\(\mathcal{M}_{EOdd}\)}}     &Unitask   &0.54 \\
      &          &Multitask  &0.76 \\
      &          &Baseline UW  	&1.31 \\
      \cdashline{3-4}[.4pt/2pt] &          
                &\methodname \textbf{(Ours)}  	&\textbf{1.17} \\

    \cline{2-4}    
    & \multirow{4}{*}{$\mathcal{M}_{EAcc}$}   &Unitask    &1.44 \\
     &           &Multitask  &0.94 \\
     &           &Baseline UW 	&1.25\\              
     \cdashline{3-4}[.4pt/2pt] &           &\methodname \textbf{(Ours)}  	&\textbf{0.95} \\
   
    \hline
  \end{tabular}
  \caption{Results for \textbf{DAIC-WOZ}. Full table results for DW, Table \ref{tab:full_daicwoz}, is available within the Appendix. Best values are highlighted in \textbf{bold}.
  }
  \label{tab:daicwoz}
  \vspace{-0.6cm}
\end{table}


\begin{figure*}[ht]
  \centerline{
  \subfigure[$\mathcal{M}_{EAcc}$ vs Acc]
  {
    \includegraphics[width=0.48\columnwidth]{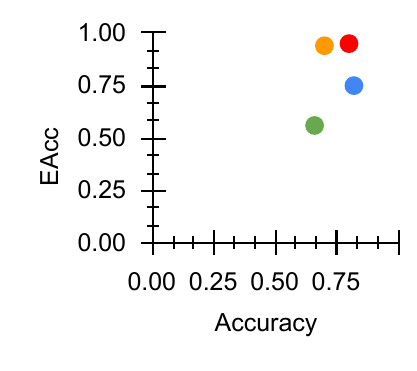}
    \label{fig:dw_EAcc}
  }
  
  \subfigure[$\mathcal{M}_{EOdd}$ vs Acc]
  {
    \includegraphics[width=0.48\columnwidth]{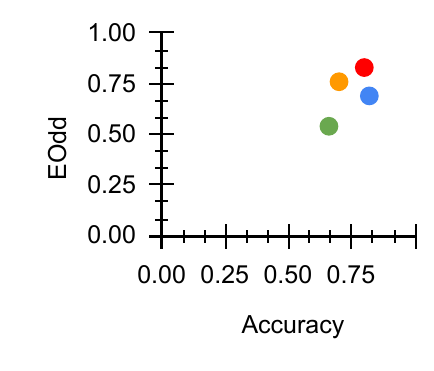}
    \label{fig:dw_EOdd}
  }
 \subfigure[$\mathcal{M}_{EOpp}$ vs Acc]
  {
    \includegraphics[width=0.48\columnwidth]{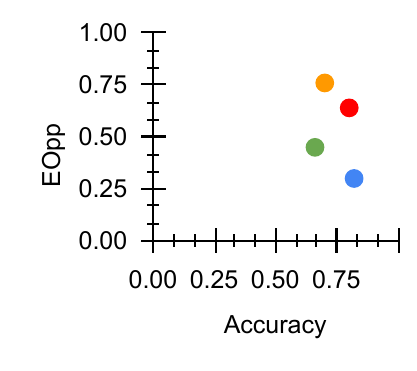}
    \label{fig:dw_EOpp}
 }
\subfigure[$\mathcal{M}_{SP}$ vs Acc]
  {
    \includegraphics[width=0.48\columnwidth]{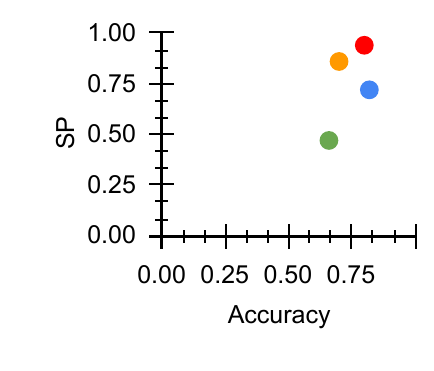}
     \label{fig:dw_SP}
 }
 }
 \vspace{-3mm}
\caption{Fairness-Accuracy Pareto Frontier across the \textbf{DAIC-WOZ} results. Upper right indicates better Pareto optimality, i.e. better fairness-accuracy trade-off. 
\textcolor{brown}{\textbf{Orange}}: Unitask. 
\textcolor{teal}{\textbf{Green}}: Multitask.
\textcolor{blue}{\textbf{Blue}}: Multitask UW.
\textcolor{purple}{\textbf{Red}}: \methodname.
Abbreviations: Acc: accuracy.
}
\label{fig:daicwoz_pareto}
\vspace{-5mm}
\end{figure*}


\begin{table}[ht!]
  \setlength{\tabcolsep}{2.5pt}
  \centering
  \small
  \begin{tabular}{l|l|l|c}
    \hline
 
   \phantom{......} & \textbf{Measure}  &\textbf{Approach}  &\textbf{Binary Outcome} 
  
   \\
    \hline


    \parbox[t]{2mm}{\multirow{20}{*}{\rotatebox[origin=c]{90}{Performance Measures}}} & \multirow{4}{*}{Acc}         &Unitask   &0.55 
    \\
      &          &Multitask   &0.58 
                \\
       &         &Baseline UW  	&0.87     \\
                \cdashline{3-4}[.4pt/2pt]
         &       &\methodname \textbf{(Ours)}    &\textbf{0.90}    \\
            
    \cline{2-4}
    & \multirow{4}{*}{F1}          &Unitask  &\textbf{0.51}    \\
       &         &Multitask  &0.45  \\
       &         &Baseline UW  &0.27    \\
                \cdashline{3-4}[.4pt/2pt]
        &        &\methodname \textbf{(Ours)} 	&0.45   \\
    \cline{2-4}
    & \multirow{4}{*}{Precision}   &Unitask  &0.36  \\
     &           &Multitask &0.32  \\
      &          &Baseline UW  	&0.28 \\
                \cdashline{3-4}[.4pt/2pt]
      &          &\methodname \textbf{(Ours)} 	&\textbf{0.46}\\
    \cline{2-4}                 
    & \multirow{4}{*}{Recall}      &Unitask   &\textbf{0.87} \\
     &           &Multitask    &0.80 \\
     &           &Baseline UW 	&0.26 \\
                \cdashline{3-4}[.4pt/2pt]
     &           &\methodname \textbf{(Ours)} 	&0.45  \\
    \cline{2-4}           
    & \multirow{4}{*}{UAR}         &Unitask  &0.63\\
      &          &Multitask &0.67  \\
      &          &Baseline UW 	&0.60 \\
                \cdashline{3-4}[.4pt/2pt]
      &          &\methodname \textbf{(Ours)}	&\textbf{0.70}\\

    \hline
    \parbox[t]{2mm}{\multirow{16}{*}{\rotatebox[origin=c]{90}{Fairness Measures}}} & \multirow{4}{*}{\textbf{\(\mathcal{M}_{SP}\)}} &Unitask  &0.65 \\
      &          &Multitask &\textbf{1.25} \\
      &          &Baseline UW 		&3.86\\
                \cdashline{3-4}[.4pt/2pt]
      &          &\methodname \textbf{(Ours)}	&1.67\\
    \cline{2-4}           
    & \multirow{4}{*}{\textbf{\(\mathcal{M}_{EOpp}\)}}  &Unitask &0.57\\
      &          &Multitask &0.81 \\
      &          &Baseline UW  	&2.31\\
                \cdashline{3-4}[.4pt/2pt]
      &          &\methodname \textbf{(Ours)} 	&\textbf{1.00 }\\
    \cline{2-4}           
     & \multirow{4}{*}{\textbf{\(\mathcal{M}_{EOdd}\)}}            &Unitask &\textbf{0.75}  \\
        &        &Multitask &1.41\\
        &        &Baseline UW  	&8.21 \\
                \cdashline{3-4}[.4pt/2pt]
        &        &\methodname \textbf{(Ours)} 	&5.00 \\
    \cline{2-4}           
    & \multirow{4}{*}{\textbf{\(\mathcal{M}_{EAcc}\)}}            &Unitask  &0.83  \\
      &          &Multitask  &0.65 \\
      &          &Baseline UW 	&0.92\\
                \cdashline{3-4}[.4pt/2pt]
      &          &\methodname \textbf{(Ours)} 	&\textbf{0.94}\\

    \hline
  \end{tabular}
  \caption{Results for \textbf{E-DAIC}. Full table results for ED, Table \ref{tab:full_E-DAIC}, is available within the Appendix. Best values are highlighted in \textbf{bold}. }
  \label{tab:E-DAIC}
  \vspace{-0.3cm}
\end{table}

\begin{figure*}[ht]
  \centerline{
  \subfigure[$\mathcal{M}_{EAcc}$ vs Acc]
  {
    \includegraphics[width=0.48\columnwidth]{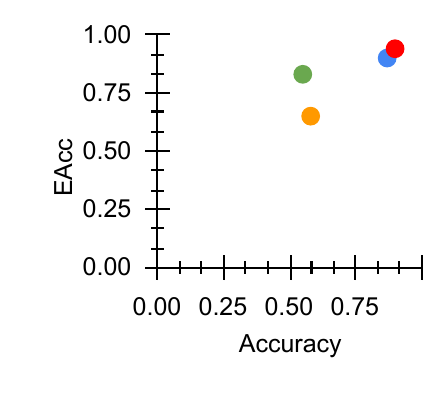}
    \label{fig:ed_EAcc}
  }
  
  \subfigure[$\mathcal{M}_{EOdd}$ vs Acc]
  {
    \includegraphics[width=0.48\columnwidth]{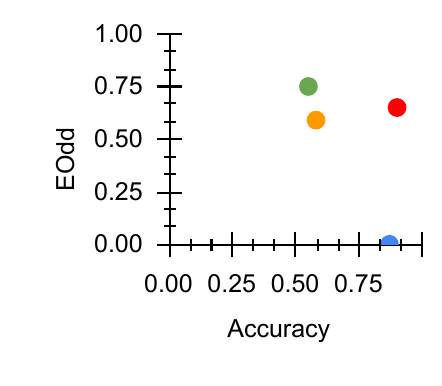}
    \label{fig:ed_EOdd}
  }
 \subfigure[$\mathcal{M}_{EOpp}$ vs Acc]
  {
    \includegraphics[width=0.48\columnwidth]{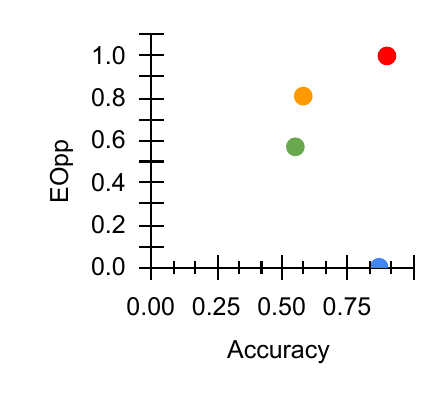}
    \label{fig:ed_EOpp}
 }
\subfigure[$\mathcal{M}_{SP}$ vs Acc]
  {
    \includegraphics[width=0.48\columnwidth]{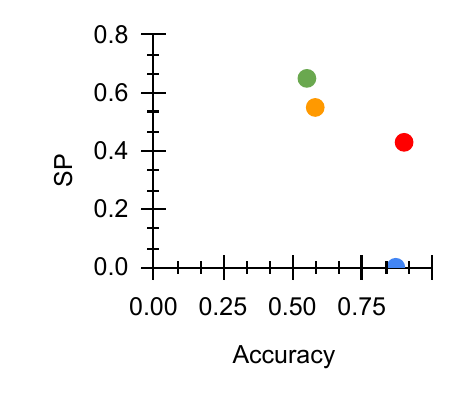}
     \label{fig:ed_SP}
 }
 }
 \vspace{-3mm}
\caption{Fairness-Accuracy Pareto Frontier across the \textbf{E-DAIC} results. Upper right indicates better Pareto optimality, i.e. better fairness-accuracy trade-off. 
\textcolor{brown}{\textbf{Orange}}: Unitask. 
\textcolor{teal}{\textbf{Green}}: Multitask.
\textcolor{blue}{\textbf{Blue}}: Multitask UW.
\textcolor{purple}{\textbf{Red}}: \methodname.
Abbreviations: Acc: accuracy.
}
\label{fig:E-DAIC_pareto}
\vspace{-5mm}
\end{figure*}

%% file: Main_Content/5_Results.tex
\section{Results}

For both datasets, we normalise the fairness results to facilitate visualisation in Figures \ref{fig:daicwoz_pareto} and \ref{fig:E-DAIC_pareto}.

\begin{table}[ht]
    \centering 
    \small
    \setlength{\tabcolsep}{4.0pt}
  \begin{tabular}{l|c|c|c}
    \hline
    \textbf{Method} & \textbf{Prec.} & \textbf{Rec.} & \textbf{F1} \\
    \hline

    \cite{ma2016depaudionet} &0.35 &\textbf{1.00} &0.52\\  
    
    \cite{song2018human} &0.32 &0.86 &0.46 \\
    
    \cite{williamson2016detecting} &- &- &0.53\\  
    
    \cite{song2018human}  &0.60 &0.43 &0.50 \\

    \cdashline{1-4}[.4pt/2pt]
    \methodname (Ours)   &0.52 &0.60 &\textbf{0.57}\\
    
    \hline
  \end{tabular}

    \caption[Test]{Comparison with other models which used extracted features for DAIC-WOZ. 
    Best results highlighted in \textbf{bold}. }
    \label{tab:comparison_extracted_features_all} 
    \vspace*{-1cm}
\end{table}


\subsection{Uni vs Multitask}
For DAIC-WOZ (DW), we see from Table \ref{tab:daicwoz},
we find that a multitask approach generally improves results compared to a unitask approach (Section \ref{sect:unitask_approach}). 
The baseline loss re-weighting approach from Equation \ref{eqn:UW_MTL} managed to further improve \textit{performance}. 
For example, 
we see from Table \ref{tab:daicwoz} that the overall classification accuracy improved from $0.70$ within a vanilla MTL approach to $0.82$ using the baseline uncertainty-based task reweighing approach. 

However, this observation is not consistent for E-DAIC (ED).
With reference to Table \ref{tab:E-DAIC}, a unitask approach seems to perform better. We see 
evidence of \textit{negative transfer}, i.e. the phenomena where learning multiple tasks concurrently result in lower performance than a unitask approach. 
We hypothesise that this is because ED is a more challenging dataset.
When adopting a multitask approach, the model completely relies on the easier tasks thus negatively impacting the learning of the other tasks. 

Moreover, performance improvement seems to come at a cost. This may be due to the fairness-accuracy trade-off \citep{wang2021understanding}. 
For instance in DW, we see that the fairness scores  
$\mathcal{M}_{SP}$, $\mathcal{M}_{EOpp}$, $\mathcal{M}_{Odd}$ and $\mathcal{M}_{Acc}$ reduced from $0.86$, $0.78$, $0.94$ and $0.76$ to $1.23$, $1.70$, $1.31$ and $1.25$ respectively. 
This is consistent with the analysis across the Pareto frontier depicted in Figures \ref{fig:daicwoz_pareto} and \ref{fig:E-DAIC_pareto}.

\subsection{Uncertainty \& the Pareto Frontier}

Our proposed loss reweighting approach seems to address the negative transfer and Pareto frontier challenges. 
Although accuracy dropped slightly from $0.82$ to $0.80$, fairness largely improved compared to the baseline UW approach (Equation \ref{eqn:UW_MTL}).
We see from Table \ref{tab:daicwoz} that fairness improved across $\mathcal{M}_{SP}$, $\mathcal{M}_{EOpp}$, $\mathcal{M}_{EOdd}$ and $\mathcal{M}_{Acc}$ from $1.23$, $1.70$, $1.31$, $1.25$ to $1.06$, $1.46$, $1.17$ and $0.95$ for DW. 
For ED, the baseline UW which adopts a task based difficulty reweighting mechanism seems to somewhat mitigate the task-based negative transfer which improves the unitask performance but not overall performance nor fairness measures. 
Our proposed method which takes into account the gender difference may have somewhat addressed this task-based negative transfer. 
In concurrence, \methodname also addressed the initial bias present. We see from Table \ref{tab:E-DAIC} that fairness improved across all fairness measures. 
The scores improved from $3.86$, $2.31$, $8.21$, $0.92$ to $1.67$, $1.00$, $5.00$ and $0.94$ across $\mathcal{M}_{SP}$, $\mathcal{M}_{EOpp}$, $\mathcal{M}_{EOdd}$ and $\mathcal{M}_{Acc}$.

The Pareto frontier across all four measures illustrated in Figures \ref{fig:daicwoz_pareto} and \ref{fig:E-DAIC_pareto} demonstrated that our proposed method generally provides better accuracy-fairness trade-off across most fairness measures for both datasets.
With reference to Figure \ref{fig:daicwoz_pareto}, we see that \methodname,
generally provides a slightly better Pareto optimality compared to other methods.
This improvement in the Pareto frontier is especially pronounced for Figure \ref{fig:E-DAIC_pareto}(c).
The difference in the Pareto frontier between our proposed method and other compared methods is greater in ED (Fig \ref{fig:E-DAIC_pareto}), the more challenging dataset, compared to that in DW  (Fig \ref{fig:daicwoz_pareto}).

\begin{figure*}[hbt!]
  \centerline{
  \subfigure[DAIC-WOZ:\textbf{Female}]
  {
    \includegraphics[height=0.35\columnwidth]{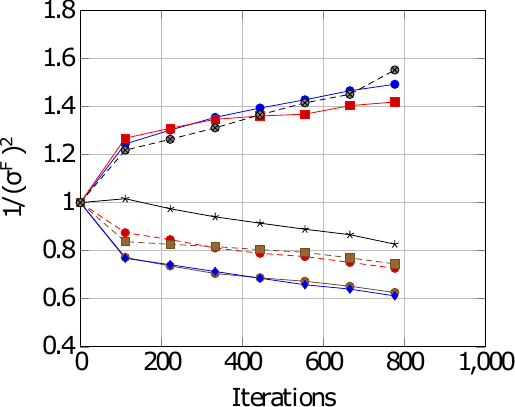}
    \label{fig:taskweights_F_daicwoz}
  }
  \subfigure[DAIC-WOZ: \textbf{Male}]
  {
    \includegraphics[height=0.35\columnwidth]{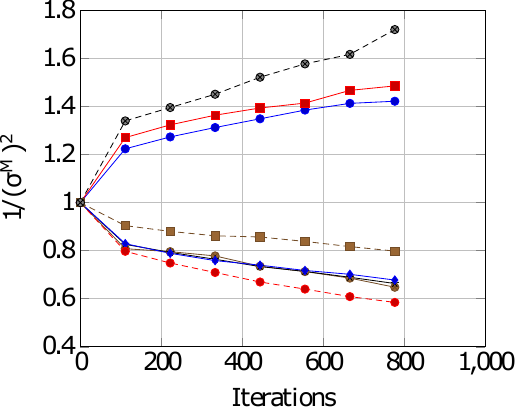}
    \label{fig:taskweights_M_daicwoz}
  }
 \subfigure[E-DAIC: \textbf{Female}]
  {
    \includegraphics[height=0.35\columnwidth]{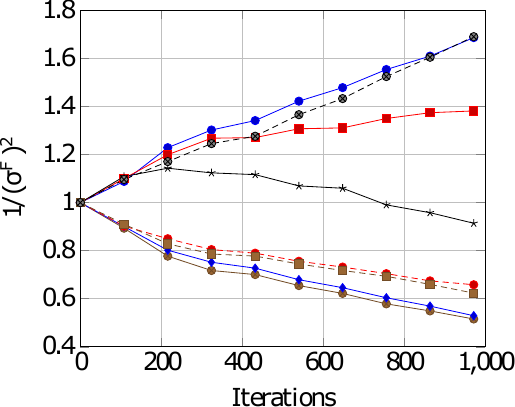}
    \label{fig:taskweights_F_edaic}
 }
\subfigure[E-DAIC: \textbf{Male}]
  {
    \includegraphics[height=0.35\columnwidth]{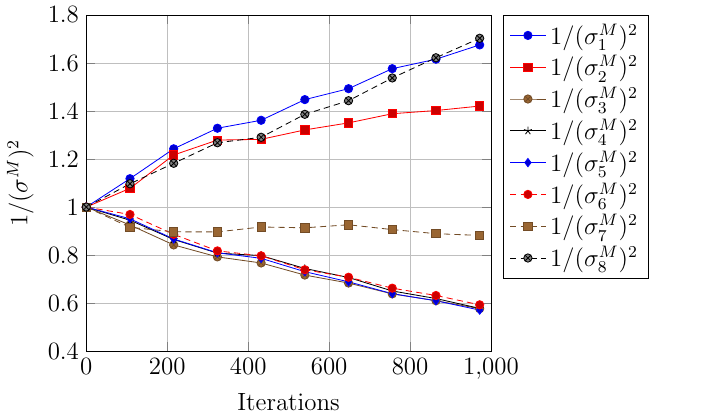}
     \label{fig:taskweights_M_edaic}
 }
 }
 \vspace{-2mm}
\caption{Task-based weightings for both gender and datasets. 
}
\label{fig:task_weights}
\vspace{-0.2cm}
\end{figure*}

For DW, with reference to 
Figures \ref{fig:taskweights_F_daicwoz} and \ref{fig:taskweights_M_daicwoz}, we see that there is a difference in task difficulty. 
Task 4 and 6 is easier for females whereas task 7 is easier for males. 
For ED, with reference to 
Figures \ref{fig:taskweights_F_edaic}, \ref{fig:taskweights_M_edaic} and Table \ref{tab:discrimination_vs_difficulty},
Task 4 seems to be easier for females whereas task 7 seems easier for males. 
Thus, adopting a gender-based uncertainty reweighting approach might have ensured that the tasks are more appropriately weighed leading towards better performance for both genders whilst mitigating the negative transfer and Pareto frontier challenges.

\vspace{-0.2cm}
\subsection{Task Difficulty \& Discrimination Capacity}

A particularly relevant and exciting finding is that each PHQ-8 subitem's task difficulty agree with its \textit{discrimination capacity} as evidenced by the rigorous study conducted by \citet{de2023reliability}. This largest study to date assessed the internal structure, reliability and cross-country validity of the PHQ-8 for the assessment of depressive symptoms.
\textit{Discrimination capacity} is defined as the ability of item to distinguish whether a person is depressed or not. 

With reference to Table \ref{tab:discrimination_vs_difficulty}, 
it is noteworthy that the task difficulty captured by 
$\frac{1}{\sigma^2}$ 
in our experiments corresponds to the discrimination capacity (DC) of each task. 
The higher $\sigma_t$, 
the more difficult the task $t$. 
In other words, the lower the value of $\frac{1}{\sigma^2}$, the more difficult the task.
For instance, in their study, PHQ-1, 2 and 6 were the items that has the greatest ability to discriminate whether a person is depressed. This is in alignment with our results where PHQ-1,2 and 8 are easier across both datasets. PHQ-3 and PHQ-5 are the least discriminatory or more difficult tasks as evidenced by the values highlighted in red.


\begin{table}[ht]
    \centering\footnotesize
    \setlength{\tabcolsep}{4.0pt}
    \begin{tabular}{l|c|c|c|c|c}
    \toprule
          & & \multicolumn{4}{c}{$\frac{1}{\sigma^2}$}\\
          \cdashline{3-6}[.4pt/2pt]
          & \multicolumn{1}{c|}{DC} & \multicolumn{1}{c|}{DW-F} & \multicolumn{1}{c|}{DW-M} & \multicolumn{1}{c|}{ED-F} & \multicolumn{1}{c}{ED-M} \\
 
    \hline
    PHQ-1 & \textcolor{teal}{\textbf{3.06}}  &\textcolor{teal}{\textbf{1.50}}      &\textcolor{teal}{\textbf{1.41}}    &\textcolor{teal}{\textbf{1.69}}   &\textcolor{teal}{\textbf{1.69}} \\
    PHQ-2 & \textcolor{teal}{\textbf{3.42}}  &\textcolor{teal}{\textbf{1.41}}   &\textcolor{teal}{\textbf{1.47}} &\textcolor{teal}{\textbf{1.38}}  &\textcolor{teal}{\textbf{1.41}} \\
    PHQ-3 & \textcolor{red}{\textbf{1.91}}   &\textcolor{red}{\textbf{0.62}}  &\textcolor{red}{\textbf{0.64}} &\textcolor{red}{\textbf{0.51}}   &\textcolor{red}{\textbf{0.58}} \\
    PHQ-4 & 2.67  &0.82    &0.68   &0.91   &0.60 \\
    PHQ-5 & \textcolor{red}{\textbf{2.22}}   &\textcolor{red}{\textbf{0.61}}  &0.69   &\textcolor{red}{\textbf{0.51}}  &\textcolor{red}{\textbf{0.58}} \\
    PHQ-6 & \textcolor{teal}{\textbf{2.86}}   &0.73     &\textcolor{red}{\textbf{0.59}}  &0.63     &0.60 \\
    PHQ-7 & 2.55  &0.75    &0.80   &0.61    &0.89 \\
     PHQ-8 & 2.43  &\textcolor{teal}{\textbf{1.58}}    &\textcolor{teal}{\textbf{1.72}} &\textcolor{teal}{\textbf{1.69}}  &\textcolor{teal}{\textbf{1.70}} \\

    \bottomrule
    \end{tabular}
    \caption{Discrimination capacity (DC) vs 
    $\frac{1}{\sigma^2}$.
    Lower $\frac{1}{\sigma^2}$ values implies higher task difficulty.
    \textcolor{teal}{\textbf{Green}}: 
    top 3 highest scores. \textcolor{red}{\textbf{Red}}:
    bottom 2 lowest scores. 
    Our results are in harmony with the largest and most comprehensive study on the PHQ-8 conducted by \citet{de2023reliability}. 
    DW: DAIC-WOZ. ED: E-DAIC. F: Female. M: Male.
    }
    \label{tab:discrimination_vs_difficulty}
   \vspace{-1cm}
\end{table}

%% file: Main_Content/6_Discussion.tex
\section{Discussion and Conclusion}

Our experiments unearthed several interesting insights. 
First, overall, there are certain gender-based differences across the different PHQ-8 distribution labels as evidenced in Figure \ref{fig:task_weights}.
In addition, each task have slightly different degree of task uncertainty across gender.
This may be due to a gender difference in PHQ-8 questionnaire profiling or inadequate data curation. 
Thus, employing a gender-aware approach may be a viable method to improve fairness and accuracy for depression detection.

Second, though a multitask approach generally performs better than a unitask approach, this comes with several caveats. 
We see from Table \ref{tab:discrimination_vs_difficulty} that each task has a different level of difficulty.
Naively using all tasks may worsen performance and fairness compared to a unitask approach if we do not account for task-based uncertainty.
This is in agreement with existing literature which indicates that there can be a mix of positive and negative transfers across tasks 
\citep{li2023identification} and tasks have to be related for performance to improve \citep{wang2021fair}. 

Third, understanding, analysing and improving upon the fairness-accuracy Pareto frontier within the task of depression requires a nuanced and careful use of measures and datasets in order to avoid the fairness-accuracy trade-off.
Moreover, there is a growing amount of research indicating that if using appropriate methodology and metrics, these trade-offs are not always present
\citep{dutta2020there,black2022model,cooper2021emergent}
and can be mitigated with careful selection of models \citep{black2022model}
and evaluation methods \citep{wick2019unlocking}.
Our results are in agreement with existing works indicating that state-of-the-art bias mitigation methods are typically only effective at removing epistemic discrimination \citep{wang2023aleatoric}, i.e. the discrimination made during model development, but not aleatoric discrimination.
In order to address aleatoric discrimination, i.e. the bias inherent within the data distribution, and to improve the Pareto frontier, better data curation is required \citep{dutta2020there}.
Though our results are unable to provide a significant improvement on the Pareto frontier, we believe that this work presents the first step in this direction and would encourage future work to look into this. 
In sum, we present a novel gender-based uncertainty multitask loss reweighting mechanism. 
We showed that our proposed multitask loss reweighting is able to improve fairness with lesser fairness-accuracy trade-off. 
Our findings also revealed the importance of accounting for 
negative transfers and for more effort to be channelled towards improving the Pareto frontier in depression detection research. 

\vspace{-0.1cm}
\paragraph{ML for Healthcare Implication:}

Producing a thorough review of strategies to improve fairness is not within the scope of this work. 
Instead, the key goal is to advance ML for healthcare solutions that are grounded in the framework used by clinicians. In our settings, this corresponds to using each PHQ-8 subcriterion as individual subtask within our MTL-based approach
and using a
a gender-based uncertainty reweighting mechanism to account for the gender difference in PHQ-8 label distribution.
By replicating the inferential process used by clinicians, this work attempts to bridge ML methods with the symptom-based profiling system used by clinicians. 
Future work can also make use of this property during inference in order to improve the trustworthiness of the machine learning or decision-making model \citep{huang2022trustsleepnet}.

In the process of doing so, our proposed method also provide the elusive first evidence that each PHQ-8 subitem's \textit{task difficulty} aligns with its \textit{discrimination capacity} as evidenced from data collected from the largest PHQ-8
population-based study to date
\citep{de2023reliability}.
We hope this piece of work will encourage other ML and healthcare researchers to further investigate methods that could bridge 
ML experimental results with empirical real world healthcare findings to ensure its reliability and validity.

\paragraph{Limitations:}
We only investigated gender fairness due to the limited availability of other sensitive attributes in both datasets.
Future work can consider investigating this approach across different sensitive attributes such as race and age, the intersectionality of sensitive attributes and other healthcare challenges such as cognitive impairment or cancer diagnosis. 
Moreover, we have adopted our existing experimental approach in alignment with the train-validation-test split provided by the dataset owners as well as other existing works. Future works can consider adopting a cross-validation approach.
Other interesting directions include investigating this challenge as an ordinal regression problem \citep{diaz2019soft}.
Future work can also consider repeating the experiments using datasets collected from other countries and dive deeper into the cultural intricacies of the different PHQ-8 subitems, investigate the effects of the different modalities and its relation to a multitask approach, as well as investigate other important topics such as interpretability and explainability to advance responsible \citep{wiens2019no} and ethical machine learning for healthcare \citep{chen2021ethical}.

%% file: Main_Content/appendix.tex

\section{Experimental Setup}
\subsection{Datasets}

For both DAIC-WOZ and E-DAIC, we 
work with the extracted features and 
followed the train-validate-test split provided. 
The dataset owners provided the ground-truths for each of the PHQ-8 sub-criterion and the final binary classification for both datasets.

\paragraph{DAIC-WOZ \citep{valstar2016avec}} contains audio recordings, extracted visual features and transcripts  
collected in a lab-based setting of 100 males and 85 females. 
The dataset owners provided a standard train-validate-test split which we followed.
The dataset owners also provided the ground-truths for each of the PHQ-8 questionnaire sub-criterion as well as the final binary classification.

\paragraph{E-DAIC 
\citep{ringeval2019avec}} contains acoustic recordings and extracted visual features of 168 males and 103 females. 
The dataset owners provided a standard train-validate-test split which we followed.
%
%


\begin{table*}[ht!]
  \setlength{\tabcolsep}{2.5pt}
  \footnotesize
  \centering
  \begin{tabular}{l|l|cccccccc|c|}
    \hline

    Measure  &Approach &PHQ-1	& PHQ-2	& PHQ-3	& PHQ-4	& PHQ-5	& PHQ-6	& PHQ-7	& PHQ-8 &  \textbf{Binary Outcome} \\


    \hline

    \multirow{4}{*}{Acc}     &Unitask        &\textbf{0.87} &0.51 &0.62 &0.57 &0.57 &0.51 &\textbf{0.79} &\textbf{0.94} &0.66    \\
            &Multitask      &0.72 &0.68 &0.57 &\textbf{0.62} &0.64 &\textbf{0.68} &0.74 &0.89 &0.70             \\
            &Baseline UW    &0.81 &\textbf{0.70}	&\textbf{0.64}	&0.60	&\textbf{0.66}	&0.62	&0.72	&0.87	&\textbf{0.82}    \\
            \cdashline{2-11}[.4pt/2pt]
            & \methodname \textbf{(Ours)}       &0.68 &0.66	 &0.47	 &0.43	 &0.43	 &0.49	 &0.60	 &0.74	 &0.80   \\

    \hline              
    \multirow{4}{*}{F1}          &Unitask             &0.25 &0.41 &0.44 &0.33 &\textbf{0.33} &\textbf{0.53} &0.44 &\textbf{0.40} &0.47 \\
                &Multitask           &0.32 &0.29 &0.50 &\textbf{0.44} &0.32 &0.48 &\textbf{0.45} &0.29 &0.53 \\
                &Baseline UW         &\textbf{0.40}	&0.30	&\textbf{0.51}	&0.42	&\textbf{0.33}	&0.31	&0.43	&0.25	&0.29    \\
                \cdashline{2-11}[.4pt/2pt]
                &\methodname \textbf{(Ours)}            &0.29	 &\textbf{0.33}	 &0.44	 &0.43	 &0.27	 &0.33	 &0.39	 &0.00	 &\textbf{0.54}   \\
                
    \hline
    \multirow{4}{*}{Precision}   &Unitask        &\textbf{1.00} &0.27 &0.47 &0.31 &0.26 &0.37 &0.67 &\textbf{0.50} &0.44 \\
                &Multitask      &0.25 &0.25 &0.43 &\textbf{0.39} &0.29 &\textbf{0.47} &\textbf{0.50} &0.25 &0.50 \\
                &Baseline UW    &0.38	&0.27	&\textbf{0.50}	&0.37	&\textbf{0.31}	&0.33	&0.45	&0.20	&0.22 \\
                \cdashline{2-11}[.4pt/2pt]
                &\methodname \textbf{(Ours)}       &0.21	 &\textbf{0.27}	 &0.36	 &0.30	 &0.19	 &0.27	 &0.32	 &0.00	 &\textbf{0.56}\\
                
    \hline
    \multirow{4}{*}{Recall}      &Unitask      &0.14 &\textbf{0.89} &0.41 &0.36 &0.45 &\textbf{0.93} &0.33 &\textbf{0.33} &0.50 \\
                &Multitask    &0.43 &0.33 &0.59 &0.50 &0.36 &0.50 &0.42 &0.33 &0.57 \\
                &Baseline UW   &0.43	&0.33	&0.53	&0.50	&0.36	&0.29	&0.42	&0.33	&0.43  \\
                \cdashline{2-11}[.4pt/2pt]
                &\methodname \textbf{(Ours)}   &\textbf{0.43}	 &0.44	 &\textbf{0.59}	 &\textbf{0.71}	 &\textbf{0.45}	 &0.43	 &\textbf{0.50}	 &0.00	 &\textbf{0.60}  \\
                
    \hline
    \multirow{4}{*}{UAR}         &Unitask  &\textbf{0.93} &\textbf{0.60} &0.58 &0.51 &0.52 &\textbf{0.64 } &\textbf{0.74} &\textbf{0.73} &0.60 \\
                &Multitask  &0.57 &0.54 &0.57 &0.57 &0.54 &0.62 &0.66 &0.60 &\textbf{0.65} \\
                &Baseline UW  &0.65	&0.56	&\textbf{0.61}	&\textbf{0.57}	&\textbf{0.56}	&0.52	&0.62	&0.62	&0.64 \\
                \cdashline{2-11}[.4pt/2pt]
                &\methodname \textbf{(Ours)}  &0.58	 &0.58	 &0.49	 &0.51	 &0.44	 &0.47	 &0.56	 &0.40	 &0.63\\
     \hline

    \multirow{4}{*}{\textbf{\(\mathcal{M}_{SP}\)}} &Unitask   &0.00 &1.44 &1.92 &1.60 &0.86 &1.44 &4.79 &\textbf{0.96} &0.47 \\
                &Multitask  &1.92 &\textbf{0.96} &1.80 &1.20 &3.51 &\textbf{1.10} &3.83 &2.88 &0.86 \\
                &Baseline UW  &2.88	&1.15	&1.92	&\textbf{1.06}	&2.16	&1.34	&1.15	&1.44	&1.23\\
                \cdashline{2-11}[.4pt/2pt]
                &\methodname \textbf{(Ours)}  &\textbf{0.72}	&0.64	&\textbf{1.28}	&1.15	&\textbf{1.12}	&0.66	&\textbf{0.86}	&0.77	&\textbf{1.06}\\
    \hline            
    \multirow{4}{*}{\textbf{\(\mathcal{M}_{EOpp}\)}}   &Unitask  &0.00 &1.50 &2.00 &1.67 &0.90 &1.50 &5.00 &1.00 &0.45 \\
                &Multitask  &2.00 &\textbf{1.00} &1.88 &1.25 &3.67 &\textbf{1.14} &4.00 &3.00 &\textbf{0.78} \\
                &Baseline UW  &3.00	&1.20	&2.00	&\textbf{1.11}	&2.25	&1.40	&1.20	&1.50	&1.70 \\
                \cdashline{2-11}[.4pt/2pt]
                &\methodname \textbf{(Ours)}  &\textbf{0.75}    &0.67	&\textbf{1.33}	&1.20	&\textbf{1.17}	&0.69	&\textbf{0.90}	&\textbf{0.80}	&1.46 \\
            
    \hline       
    \multirow{4}{*}{\textbf{\(\mathcal{M}_{EOdd}\)}}     &Unitask  &0.00 &1.44 &1.90 &2.83 &1.25 &1.53 &0.00 &0.00 &0.54 \\
                &Multitask  &0.00 &1.60 &1.83 &1.28 &9.00 &1.88 &4.00 &0.00 &0.76 \\
                &Baseline UW  &0.00	&0.00	&2.29	&1.49	&3.50	&2.25	&1.50	&2.74	&1.31 \\
                \cdashline{2-11}[.4pt/2pt]
                &\methodname \textbf{(Ours)}  &\textbf{0.80}	&\textbf{0.80}	&\textbf{1.43}	&\textbf{1.16}	&\textbf{1.33}	&\textbf{0.75}	&\textbf{1.00}	&0.00	&\textbf{1.17} \\

    \hline    
    \multirow{4}{*}{$\mathcal{M}_{EAcc}$}   &Unitask  &0.91 &0.81 &\textbf{0.89} &0.56 &\textbf{1.20} &0.81 &\textbf{1.01} &\textbf{0.96} &1.44 \\
                &Multitask &0.96 &\textbf{1.09} &0.89 &0.89 &0.55 &1.23 &1.01 &0.87 &0.94 \\
                &Baseline UW &\textbf{0.96}	&1.30	&0.84	&0.72	&0.69	&\textbf{1.03}	&1.08	&0.91	&1.25\\
                \cdashline{2-11}[.4pt/2pt]
                &\methodname \textbf{(Ours)}  &1.09	&1.16	&0.80	&\textbf{0.96}	&0.64	&1.28	&1.11	&1.14	&\textbf{0.95} \\
   
    \hline
  \end{tabular}
  \caption{Full experimental results for \textbf{DAIC-WOZ} across the different PHQ-8 subitems.
  Best values are highlighted in \textbf{bold}.
  }
  \label{tab:full_daicwoz}
\end{table*}


\begin{table*}[ht!]
  \setlength{\tabcolsep}{2.5pt}
  \centering
  \footnotesize
  \begin{tabular}{l|l|cccccccc|c|}
    \hline
 
   Measure  &Approach &PHQ-1	& PHQ-2	& PHQ-3	& PHQ-4	& PHQ-5	& PHQ-6	& PHQ-7	& PHQ-8 &  \textbf{Binary Outcome} 
  
   \\
    \hline


    \multirow{4}{*}{Acc}         &Unitask  &\textbf{0.80} & \textbf{0.66} & 0.59 & 0.66 & 0.59 & 0.61 & 0.63 & 0.89 &0.55 
    \\
                &Multitask   &0.68 &0.54 &0.48 &0.43 &0.52 &0.54 &0.48 &0.54 &0.58 
                \\
                &Baseline UW  &0.75	&0.63	&\textbf{0.61}	&\textbf{0.73}	&\textbf{0.73}	&0.63	&0.59	&0.89	&0.87     \\
                \cdashline{2-11}[.4pt/2pt]
                &\methodname \textbf{(Ours)}   &0.77	&0.61	&0.61	&0.54	&0.71	&\textbf{0.71}	&\textbf{0.71}	&\textbf{0.93} &\textbf{0.90}    \\
            
    \hline
    \multirow{4}{*}{F1}          &Unitask & \textbf{0.27} & 0.24 &0.49 &\textbf{0.60} &\textbf{0.47} &\textbf{0.45} & \textbf{0.49} & \textbf{0.25} &\textbf{0.51}    \\
                &Multitask  &0.18 &0.32 &0.47 &0.43 &0.40 &0.38 &0.38 &0.07 &0.45  \\
                &Baseline UW  &0.22	&\textbf{0.36}	&\textbf{0.54}	&0.48	&0.29	&0.09	&0.08	&0.00	&0.27    \\
                \cdashline{2-11}[.4pt/2pt]
                &\methodname \textbf{(Ours)}  & 0.13	&0.21	&0.39	&0.43	&0.33	&0.33	&0.27	&0.00	&0.45   \\
    \hline
    \multirow{4}{*}{Precision}   &Unitask & \textbf{0.29} & 0.21 &0.38 &{0.45} & 0.34 & 0.33 &\textbf{0.33} &\textbf{0.25} &0.36  \\
                &Multitask &0.14 &0.22 &0.33 &0.30 &0.29 &0.28 &0.25 &0.04 &0.32  \\
                &Baseline UW   &0.20	&\textbf{0.27}	&\textbf{0.41}	&\textbf{0.54}	&\textbf{0.43}	&0.10	&0.07	&0.00	&0.28 \\
                \cdashline{2-11}[.4pt/2pt]
                &\methodname \textbf{(Ours)} &0.14	&0.18	&0.35	&0.33	&0.40	&\textbf{0.36}	&0.27	&0.00	&\textbf{0.46}\\
    \hline                 
    \multirow{4}{*}{Recall}      &Unitask  & \textbf{0.25} & 0.27 & 0.69 &\textbf{0.88} &\textbf{0.71} &\textbf{0.69} &\textbf{0.91} & \textbf{0.25} &\textbf{0.87} \\
                &Multitask    &0.25 &\textbf{0.55} &\textbf{0.81} &0.75 &0.64 &0.62 &0.82 &0.25 &0.80 \\
                &Baseline UW  &0.25	&0.55	&0.81	&0.44	&0.21	&0.08	&0.09	&0.00	&0.26 \\
                \cdashline{2-11}[.4pt/2pt]
                &\methodname \textbf{(Ours)} & 0.13	&0.27	&0.44	&0.63	&0.29	&0.31	&0.27	&0.00	&0.45  \\
    \hline           
    \multirow{4}{*}{UAR}         &Unitask &0\textbf{.58} & 0.51 & 0.60 &\textbf{0.69} &\textbf{0.60} &\textbf{0.60} &\textbf{0.65} & \textbf{0.60} &0.63\\
                &Multitask &0.50 &0.52 &0.58 &0.53 &0.55 &0.55 &0.58 &0.47 &0.67  \\
                &Baseline UW &0.54	&\textbf{0.59}	&\textbf{0.67}	&0.64	&0.56	&0.43	&0.40	&0.48	&0.60 \\
                \cdashline{2-11}[.4pt/2pt]
                &\methodname \textbf{(Ours)} &0.50	&0.48	&0.56	&0.56	&0.57	&0.57	&0.55	&0.50	&\textbf{0.70}\\

    \hline
    \multirow{4}{*}{\textbf{\(\mathcal{M}_{SP}\)}} &Unitask  & 0.26 & 2.78 & 0.81 &\textbf{1.12} & 0.94 & 1.44 &\textbf{1.03} &\textbf{0.52} &0.65 \\
                &Multitask &5.67 &2.63 &\textbf{1.19} &1.40 &\textbf{0.98} &1.44 &1.24 &0.41 &\textbf{1.25} \\
                &Baseline UW  &\textbf{1.55}	&\textbf{1.29}	&2.58	&2.47	&2.06	&2.32	&5.67	&0.00	&3.86\\
                \cdashline{2-11}[.4pt/2pt]
                &\methodname \textbf{(Ours)} &2.06	&2.83	&1.26	&2.67	&3.61	&\textbf{1.29}	&1.29	&0.00	&1.67\\
    \hline           
    \multirow{4}{*}{\textbf{\(\mathcal{M}_{EOpp}\)}}  &Unitask & 0.17 & 1.80 & 0.53 & 0.72 & 0.61 &\textbf{0.93} & 0.67 & 0.33 &0.57\\
                &Multitask &3.67 &1.70 &0.77 &\textbf{0.90} &0.63 &0.93 &0.80 &0.26 &0.81 \\
                &Baseline UW  &\textbf{1.00}	&\textbf{0.83}	&1.67	&1.60	&\textbf{1.33}	&1.50	&3.67	&0.00	&2.31\\
                \cdashline{2-11}[.4pt/2pt]
                &\methodname \textbf{(Ours)} &1.33	&1.83	&\textbf{0.82}	&1.73	&2.33	&0.83	&\textbf{0.83}	&0.00	&\textbf{1.00 }\\
    \hline           
    \multirow{4}{*}{\textbf{\(\mathcal{M}_{EOdd}\)}}            &Unitask & 0.35 & 3.65 & 1.39 & 1.38 &\textbf{1.00} & \textbf{1.46} & \textbf{1.40} & \textbf{0.74} &\textbf{0.75}  \\
                &Multitask &7.00 &3.42 &\textbf{1.29} &\textbf{1.63} &1.03 &1.53 &1.43 &0.41 &1.41\\
                &Baseline UW  &3.00	&\textbf{1.76}	&4.20	&6.11	&2.00	&0.00	&0.00	&0.00	&8.21 \\
                \cdashline{2-11}[.4pt/2pt]
                &\methodname \textbf{(Ours)} &\textbf{2.80}	&3.42	&2.22	&3.67	&3.60	&2.25	&1.90	&0.00	&5.00 \\
    \hline           
    \multirow{4}{*}{\textbf{\(\mathcal{M}_{EAcc}\)}}            &Unitask & 1.13 & \textbf{0.74} & 1.45 & 0.84 & 1.14 &\textbf{0.96} & 0.71 & 1.08 &0.83  \\
                &Multitask  &0.63 &0.39 &0.77 &0.41 &\textbf{0.94} &0.77 &0.54 &1.77 &0.65 \\
                &Baseline UW &1.05	&0.71	&0.48	&\textbf{0.99}	&0.89	&0.81	&0.88	&1.12	&0.92\\
                \cdashline{2-11}[.4pt/2pt]
                &\methodname \textbf{(Ours)} &\textbf{0.96}	&0.64	&\textbf{1.22}	&0.47	&0.83	&0.74	&\textbf{1.03}	&\textbf{1.05}	&\textbf{0.94}\\

    \hline
  \end{tabular}
  \caption{Full experimental results for \textbf{E-DAIC} across the different PHQ-8 subitems.
  Best values are highlighted in \textbf{bold}. }
  \label{tab:full_E-DAIC}
\end{table*}

%% file: main.bbl
\begin{thebibliography}{75}
\providecommand{\natexlab}[1]{#1}
\providecommand{\url}[1]{\texttt{#1}}
\expandafter\ifx\csname urlstyle\endcsname\relax
  \providecommand{\doi}[1]{doi: #1}\else
  \providecommand{\doi}{doi: \begingroup \urlstyle{rm}\Url}\fi

\bibitem[Al~Hanai et~al.(2018)Al~Hanai, Ghassemi, and Glass]{al2018detecting}
Tuka Al~Hanai, Mohammad~M Ghassemi, and James~R Glass.
\newblock Detecting depression with audio/text sequence modeling of interviews.
\newblock In \emph{Interspeech}, pages 1716--1720, 2018.

\bibitem[Bailey and Plumbley(2021)]{bailey2021gender}
Andrew Bailey and Mark~D Plumbley.
\newblock Gender bias in depression detection using audio features.
\newblock \emph{EUSIPCO 2021}, 2021.

\bibitem[Baltaci et~al.(2023)Baltaci, Oksuz, Kuzucu, Tezoren, Konar, Ozkan, Akbas, and Kalkan]{baltaci2023class}
Zeynep~Sonat Baltaci, Kemal Oksuz, Selim Kuzucu, Kivanc Tezoren, Berkin~Kerim Konar, Alpay Ozkan, Emre Akbas, and Sinan Kalkan.
\newblock Class uncertainty: A measure to mitigate class imbalance.
\newblock \emph{arXiv preprint arXiv:2311.14090}, 2023.

\bibitem[Ban and Ji(2024)]{ban2024fair}
Hao Ban and Kaiyi Ji.
\newblock Fair resource allocation in multi-task learning.
\newblock \emph{arXiv preprint arXiv:2402.15638}, 2024.

\bibitem[Barocas et~al.(2017)Barocas, Hardt, and Narayanan]{barocas2017fairness}
Solon Barocas, Moritz Hardt, and Arvind Narayanan.
\newblock Fairness in machine learning.
\newblock \emph{NeurIPS Tutorial}, 1:\penalty0 2, 2017.

\bibitem[Barsky et~al.(2001)Barsky, Peekna, and Borus]{barsky2001somatic}
Arthur~J Barsky, Heli~M Peekna, and Jonathan~F Borus.
\newblock Somatic symptom reporting in women and men.
\newblock \emph{Journal of general internal medicine}, 16\penalty0 (4):\penalty0 266--275, 2001.

\bibitem[Black et~al.(2022)Black, Raghavan, and Barocas]{black2022model}
Emily Black, Manish Raghavan, and Solon Barocas.
\newblock Model multiplicity: Opportunities, concerns, and solutions.
\newblock In \emph{Proceedings of the 2022 ACM Conference on Fairness, Accountability, and Transparency}, pages 850--863, 2022.

\bibitem[Buolamwini and Gebru(2018)]{buolamwini2018gender}
Joy Buolamwini and Timnit Gebru.
\newblock Gender shades: Intersectional accuracy disparities in commercial gender classification.
\newblock In \emph{FAccT}, pages 77--91. PMLR, 2018.

\bibitem[Cameron et~al.(2024)Cameron, Cheong, Spitale, and Gunes]{cameron2024multimodal}
Joseph Cameron, Jiaee Cheong, Micol Spitale, and Hatice Gunes.
\newblock Multimodal gender fairness in depression prediction: Insights on data from the usa \& china.
\newblock \emph{arXiv preprint arXiv:2408.04026}, 2024.

\bibitem[Cetinkaya et~al.(2024)Cetinkaya, Kalkan, and Akbas]{cetinkaya2024ranked}
Bedrettin Cetinkaya, Sinan Kalkan, and Emre Akbas.
\newblock Ranked: Addressing imbalance and uncertainty in edge detection using ranking-based losses.
\newblock In \emph{Proceedings of the IEEE/CVF Conference on Computer Vision and Pattern Recognition}, pages 3239--3249, 2024.

\bibitem[Chen et~al.(2021)Chen, Pierson, Rose, Joshi, Ferryman, and Ghassemi]{chen2021ethical}
Irene~Y Chen, Emma Pierson, Sherri Rose, Shalmali Joshi, Kadija Ferryman, and Marzyeh Ghassemi.
\newblock Ethical machine learning in healthcare.
\newblock \emph{Annual review of biomedical data science}, 4\penalty0 (1):\penalty0 123--144, 2021.

\bibitem[Cheong et~al.(2021)Cheong, Kalkan, and Gunes]{cheong2021hitchhiker}
Jiaee Cheong, Sinan Kalkan, and Hatice Gunes.
\newblock The hitchhiker’s guide to bias and fairness in facial affective signal processing: Overview and techniques.
\newblock \emph{IEEE Signal Processing Magazine}, 38\penalty0 (6), 2021.

\bibitem[Cheong et~al.(2022)Cheong, Kalkan, and Gunes]{cheong2022counterfactual}
Jiaee Cheong, Sinan Kalkan, and Hatice Gunes.
\newblock Counterfactual fairness for facial expression recognition.
\newblock In \emph{European Conference on Computer Vision}, pages 245--261. Springer, 2022.

\bibitem[Cheong et~al.(2023{\natexlab{a}})Cheong, Kalkan, and Gunes]{cheong2023causal}
Jiaee Cheong, Sinan Kalkan, and Hatice Gunes.
\newblock Causal structure learning of bias for fair affect recognition.
\newblock In \emph{Proceedings of the IEEE/CVF Winter Conference on Applications of Computer Vision}, pages 340--349, 2023{\natexlab{a}}.

\bibitem[Cheong et~al.(2023{\natexlab{b}})Cheong, Kuzucu, Kalkan, and Gunes]{cheong_gender_fairness}
Jiaee Cheong, Selim Kuzucu, Sinan Kalkan, and Hatice Gunes.
\newblock Towards gender fairness for mental health prediction.
\newblock In \emph{IJCAI 2023}, pages 5932--5940, US, 2023{\natexlab{b}}. IJCAI.

\bibitem[Cheong et~al.(2023{\natexlab{c}})Cheong, Spitale, and Gunes]{cheong_acii_23}
Jiaee Cheong, Micol Spitale, and Hatice Gunes.
\newblock “it’s not fair!” – fairness for a small dataset of multi-modal dyadic mental well-being coaching.
\newblock In \emph{ACII}, pages 1--8, USA, sep 2023{\natexlab{c}}.

\bibitem[Cheong et~al.(2024{\natexlab{a}})Cheong, Kalkan, and Gunes]{cheong_fairrefuse}
Jiaee Cheong, Sinan Kalkan, and Hatice Gunes.
\newblock Fairrefuse: Referee-guided fusion for multi-modal causal fairness in depression detection.
\newblock In \emph{Proceedings of the Thirty-Third International Joint Conference on Artificial Intelligence, {IJCAI-24}}, pages 7224--7232, 8 2024{\natexlab{a}}.
\newblock AI for Good.

\bibitem[Cheong et~al.(2024{\natexlab{b}})Cheong, Spitale, and Gunes]{cheong2024_tac}
Jiaee Cheong, Micol Spitale, and Hatice Gunes.
\newblock Small but fair! fairness for multimodal human-human and robot-human mental wellbeing coaching, 2024{\natexlab{b}}.

\bibitem[Chua et~al.(2023)Chua, Kim, Choi, Lee, Deshpande, Schwab, Lev, Gonzalez, Gee, and Do]{chua2023tackling}
Michelle Chua, Doyun Kim, Jongmun Choi, Nahyoung~G Lee, Vikram Deshpande, Joseph Schwab, Michael~H Lev, Ramon~G Gonzalez, Michael~S Gee, and Synho Do.
\newblock Tackling prediction uncertainty in machine learning for healthcare.
\newblock \emph{Nature Biomedical Engineering}, 7\penalty0 (6):\penalty0 711--718, 2023.

\bibitem[Cooper et~al.(2021)Cooper, Abrams, and Na]{cooper2021emergent}
A~Feder Cooper, Ellen Abrams, and Na~Na.
\newblock Emergent unfairness in algorithmic fairness-accuracy trade-off research.
\newblock In \emph{Proceedings of the 2021 AAAI/ACM Conference on AI, Ethics, and Society}, pages 46--54, 2021.

\bibitem[de~la Torre et~al.(2023)de~la Torre, Vilagut, Ronaldson, Valderas, Bakolis, Dregan, Molina, Navarro-Mateu, P{\'e}rez, Bartoll-Roca, et~al.]{de2023reliability}
Jorge~Arias de~la Torre, Gemma Vilagut, Amy Ronaldson, Jose~M Valderas, Ioannis Bakolis, Alex Dregan, Antonio~J Molina, Fernando Navarro-Mateu, Katherine P{\'e}rez, Xavier Bartoll-Roca, et~al.
\newblock Reliability and cross-country equivalence of the 8-item version of the patient health questionnaire (phq-8) for the assessment of depression: results from 27 countries in europe.
\newblock \emph{The Lancet Regional Health--Europe}, 31, 2023.

\bibitem[Diaz and Marathe(2019)]{diaz2019soft}
Raul Diaz and Amit Marathe.
\newblock Soft labels for ordinal regression.
\newblock In \emph{Proceedings of the IEEE/CVF conference on computer vision and pattern recognition}, pages 4738--4747, 2019.

\bibitem[Dutta et~al.(2020)Dutta, Wei, Yueksel, Chen, Liu, and Varshney]{dutta2020there}
Sanghamitra Dutta, Dennis Wei, Hazar Yueksel, Pin-Yu Chen, Sijia Liu, and Kush Varshney.
\newblock Is there a trade-off between fairness and accuracy? a perspective using mismatched hypothesis testing.
\newblock In \emph{International conference on machine learning}, pages 2803--2813. PMLR, 2020.

\bibitem[Gal(2016)]{gal_uncertainty}
Yarin Gal.
\newblock Uncertainty in deep learning.
\newblock 2016.

\bibitem[Ghosh et~al.(2022)Ghosh, Ekbal, and Bhattacharyya]{ghosh2022multitask}
Soumitra Ghosh, Asif Ekbal, and Pushpak Bhattacharyya.
\newblock A multitask framework to detect depression, sentiment and multi-label emotion from suicide notes.
\newblock \emph{Cognitive Computation}, 14\penalty0 (1), 2022.

\bibitem[Gong and Poellabauer(2017)]{gong2017topic}
Yuan Gong and Christian Poellabauer.
\newblock Topic modeling based multi-modal depression detection.
\newblock In \emph{Proceedings of the 7th annual workshop on Audio/Visual emotion challenge}, pages 69--76, 2017.

\bibitem[Grote and Keeling(2022)]{grote2022enabling}
Thomas Grote and Geoff Keeling.
\newblock Enabling fairness in healthcare through machine learning.
\newblock \emph{Ethics and Information Technology}, 24\penalty0 (3):\penalty0 39, 2022.

\bibitem[Gupta et~al.(2023)Gupta, Singh, and Ranjan]{gupta2023multimodal}
Shelley Gupta, Archana Singh, and Jayanthi Ranjan.
\newblock Multimodal, multiview and multitasking depression detection framework endorsed with auxiliary sentiment polarity and emotion detection.
\newblock \emph{International Journal of System Assurance Engineering and Management}, 14\penalty0 (Suppl 1), 2023.

\bibitem[Hall et~al.(2022)Hall, van~der Maaten, Gustafson, Jones, and Adcock]{hall2022systematic}
Melissa Hall, Laurens van~der Maaten, Laura Gustafson, Maxwell Jones, and Aaron Adcock.
\newblock A systematic study of bias amplification.
\newblock \emph{arXiv preprint arXiv:2201.11706}, 2022.

\bibitem[Han et~al.(2024)Han, Canli, Shah, Zhang, Dino, and Kalkan]{han2024perspectives}
Mengjie Han, Ilkim Canli, Juveria Shah, Xingxing Zhang, Ipek~Gursel Dino, and Sinan Kalkan.
\newblock Perspectives of machine learning and natural language processing on characterizing positive energy districts.
\newblock \emph{Buildings}, 14\penalty0 (2):\penalty0 371, 2024.

\bibitem[He et~al.(2022)He, Niu, Tiwari, Marttinen, Su, Jiang, Guo, Wang, Ding, Wang, et~al.]{he2022deep}
Lang He, Mingyue Niu, Prayag Tiwari, Pekka Marttinen, Rui Su, Jiewei Jiang, Chenguang Guo, Hongyu Wang, Songtao Ding, Zhongmin Wang, et~al.
\newblock Deep learning for depression recognition with audiovisual cues: A review.
\newblock \emph{Information Fusion}, 80:\penalty0 56--86, 2022.

\bibitem[Hort et~al.(2022)Hort, Chen, Zhang, Sarro, and Harman]{hort2022bias}
Max Hort, Zhenpeng Chen, Jie~M Zhang, Federica Sarro, and Mark Harman.
\newblock Bias mitigation for machine learning classifiers: A comprehensive survey.
\newblock \emph{arXiv preprint arXiv:2207.07068}, 2022.

\bibitem[Huang and Ma(2022)]{huang2022trustsleepnet}
Guanjie Huang and Fenglong Ma.
\newblock Trustsleepnet: A trustable deep multimodal network for sleep stage classification.
\newblock In \emph{2022 IEEE-EMBS International Conference on Biomedical and Health Informatics (BHI)}, pages 01--04. IEEE, 2022.

\bibitem[Jansz et~al.(2000)]{jansz2000masculine}
Jeroen Jansz et~al.
\newblock Masculine identity and restrictive emotionality.
\newblock \emph{Gender and emotion: Social psychological perspectives}, pages 166--186, 2000.

\bibitem[Kaiser et~al.(2022)Kaiser, Kern, and Rügamer]{kaiser2022uncertainty}
Patrick Kaiser, Christoph Kern, and David Rügamer.
\newblock Uncertainty-aware predictive modeling for fair data-driven decisions, 2022.

\bibitem[Kendall et~al.(2018)Kendall, Gal, and Cipolla]{kendall2018multi}
Alex Kendall, Yarin Gal, and Roberto Cipolla.
\newblock Multi-task learning using uncertainty to weigh losses for scene geometry and semantics.
\newblock In \emph{CVPR}, pages 7482--7491, 2018.

\bibitem[Kingma and Ba(2014)]{kingma2014adam}
Diederik~P Kingma and Jimmy Ba.
\newblock Adam: A method for stochastic optimization.
\newblock \emph{ICLR}, 2014.

\bibitem[Kroenke et~al.(2009)Kroenke, Strine, Spitzer, Williams, Berry, and Mokdad]{kroenke2009phq}
Kurt Kroenke, Tara~W Strine, Robert~L Spitzer, Janet~BW Williams, Joyce~T Berry, and Ali~H Mokdad.
\newblock The phq-8 as a measure of current depression in the general population.
\newblock \emph{Journal of affective disorders}, 114\penalty0 (1-3):\penalty0 163--173, 2009.

\bibitem[Kuzucu et~al.(2024)Kuzucu, Cheong, Gunes, and Kalkan]{kuzucu2024uncertainty}
Selim Kuzucu, Jiaee Cheong, Hatice Gunes, and Sinan Kalkan.
\newblock Uncertainty as a fairness measure.
\newblock \emph{Journal of Artificial Intelligence Research}, 81:\penalty0 307--335, 2024.

\bibitem[Leung et~al.(2020)Leung, Mak, Leung, Chiang, and Loke]{leung2020measurement}
Doris~YP Leung, Yim~Wah Mak, Sau~Fong Leung, Vico~CL Chiang, and Alice~Yuen Loke.
\newblock Measurement invariances of the phq-9 across gender and age groups in chinese adolescents.
\newblock \emph{Asia-Pacific Psychiatry}, 12\penalty0 (3):\penalty0 e12381, 2020.

\bibitem[Li et~al.(2023{\natexlab{a}})Li, Ding, Zou, Hu, Jiang, and Zhang]{li2023multi}
Can Li, Sirui Ding, Na~Zou, Xia Hu, Xiaoqian Jiang, and Kai Zhang.
\newblock Multi-task learning with dynamic re-weighting to achieve fairness in healthcare predictive modeling.
\newblock \emph{Journal of Biomedical Informatics}, 143:\penalty0 104399, 2023{\natexlab{a}}.

\bibitem[Li et~al.(2023{\natexlab{b}})Li, Lai, Jiang, and Zhang]{li2023feri}
Can Li, Dejian Lai, Xiaoqian Jiang, and Kai Zhang.
\newblock Feri: A multitask-based fairness achieving algorithm with applications to fair organ transplantation.
\newblock \emph{arXiv preprint arXiv:2310.13820}, 2023{\natexlab{b}}.

\bibitem[Li et~al.(2024)Li, Jiang, and Zhang]{li2024transformer}
Can Li, Xiaoqian Jiang, and Kai Zhang.
\newblock A transformer-based deep learning approach for fairly predicting post-liver transplant risk factors.
\newblock \emph{Journal of Biomedical Informatics}, 149:\penalty0 104545, 2024.

\bibitem[Li et~al.(2022)Li, Braud, and Amblard]{li2022multi}
Chuyuan Li, Chlo{\'e} Braud, and Maxime Amblard.
\newblock Multi-task learning for depression detection in dialogs.
\newblock \emph{arXiv preprint arXiv:2208.10250}, 2022.

\bibitem[Li et~al.(2023{\natexlab{c}})Li, Nguyen, and Zhang]{li2023identification}
Dongyue Li, Huy Nguyen, and Hongyang~Ryan Zhang.
\newblock Identification of negative transfers in multitask learning using surrogate models.
\newblock \emph{Transactions on Machine Learning Research}, 2023{\natexlab{c}}.

\bibitem[Long et~al.(2022)Long, Lei, Peng, Xu, and Mao]{long2022scoping}
Nannan Long, Yongxiang Lei, Lianhua Peng, Ping Xu, and Ping Mao.
\newblock A scoping review on monitoring mental health using smart wearable devices.
\newblock \emph{Mathematical Biosciences and Engineering}, 19\penalty0 (8), 2022.

\bibitem[Ma et~al.(2016)Ma, Yang, Chen, Huang, and Wang]{ma2016depaudionet}
Xingchen Ma, Hongyu Yang, Qiang Chen, Di~Huang, and Yunhong Wang.
\newblock Depaudionet: An efficient deep model for audio based depression classification.
\newblock In \emph{6th Intl. Workshop on audio/visual emotion challenge}, 2016.

\bibitem[Mehta et~al.(2023)Mehta, Shui, and Arbel]{mehta2023evaluating}
Raghav Mehta, Changjian Shui, and Tal Arbel.
\newblock Evaluating the fairness of deep learning uncertainty estimates in medical image analysis, 2023.

\bibitem[Naik et~al.(2024)Naik, Kalkan, and Kr{\"u}ger]{naik2024pre}
Lakshadeep Naik, Sinan Kalkan, and Norbert Kr{\"u}ger.
\newblock Pre-grasp approaching on mobile robots: a pre-active layered approach.
\newblock \emph{IEEE Robotics and Automation Letters}, 2024.

\bibitem[Ogrodniczuk and Oliffe(2011)]{ogrodniczuk2011men}
John~S Ogrodniczuk and John~L Oliffe.
\newblock Men and depression.
\newblock \emph{Canadian Family Physician}, 57\penalty0 (2):\penalty0 153--155, 2011.

\bibitem[Pleiss et~al.(2017)Pleiss, Raghavan, Wu, Kleinberg, and Weinberger]{pleiss2017fairness}
Geoff Pleiss, Manish Raghavan, Felix Wu, Jon Kleinberg, and Kilian~Q Weinberger.
\newblock On fairness and calibration.
\newblock \emph{NeurIPS}, 30, 2017.

\bibitem[Ringeval et~al.(2019)Ringeval, Schuller, Valstar, Cummins, Cowie, and Pantic]{ringeval2019avec}
Fabien Ringeval, Bj{\"o}rn Schuller, Michel Valstar, Nicholas Cummins, Roddy Cowie, and Maja Pantic.
\newblock Avec'19: Audio/visual emotion challenge and workshop.
\newblock In \emph{ICMI}, pages 2718--2719, 2019.

\bibitem[Sendak et~al.(2020)Sendak, Elish, Gao, Futoma, Ratliff, Nichols, Bedoya, Balu, and O'Brien]{sendak2020human}
Mark Sendak, Madeleine~Clare Elish, Michael Gao, Joseph Futoma, William Ratliff, Marshall Nichols, Armando Bedoya, Suresh Balu, and Cara O'Brien.
\newblock "the human body is a black box" supporting clinical decision-making with deep learning.
\newblock In \emph{FAccT}, pages 99--109, 2020.

\bibitem[Song et~al.(2018)Song, Shen, and Valstar]{song2018human}
Siyang Song, Linlin Shen, and Michel Valstar.
\newblock Human behaviour-based automatic depression analysis using hand-crafted statistics and deep learned spectral features.
\newblock In \emph{FG 2018}, pages 158--165. IEEE, 2018.

\bibitem[Spitale et~al.(2024)Spitale, Cheong, and Gunes]{spitale2024underneath}
Micol Spitale, Jiaee Cheong, and Hatice Gunes.
\newblock Underneath the numbers: Quantitative and qualitative gender fairness in llms for depression prediction.
\newblock \emph{arXiv preprint arXiv:2406.08183}, 2024.

\bibitem[Tahir et~al.(2023)Tahir, Cheng, and Liu]{tahir2023fairness}
Anique Tahir, Lu~Cheng, and Huan Liu.
\newblock Fairness through aleatoric uncertainty.
\newblock In \emph{CIKM}, 2023.

\bibitem[Thibodeau and Asmundson(2014)]{thibodeau2014phq}
Michel~A Thibodeau and Gordon~JG Asmundson.
\newblock The phq-9 assesses depression similarly in men and women from the general population.
\newblock \emph{Personality and Individual Differences}, 56:\penalty0 149--153, 2014.

\bibitem[Valstar et~al.(2016)Valstar, Gratch, Schuller, Ringeval, Lalanne, Torres~Torres, Scherer, Stratou, Cowie, and Pantic]{valstar2016avec}
Michel Valstar, Jonathan Gratch, Bj{\"o}rn Schuller, Fabien Ringeval, Denis Lalanne, Mercedes Torres~Torres, Stefan Scherer, Giota Stratou, Roddy Cowie, and Maja Pantic.
\newblock Avec 2016: Depression, mood, and emotion recognition workshop and challenge.
\newblock pages 3--10, 2016.

\bibitem[Vetter et~al.(2013)Vetter, Wadden, Vinnard, Moore, Khan, Volger, Sarwer, and Faulconbridge]{vetter2013gender}
Marion~L Vetter, Thomas~A Wadden, Christopher Vinnard, Rene{\'e}~H Moore, Zahra Khan, Sheri Volger, David~B Sarwer, and Lucy~F Faulconbridge.
\newblock Gender differences in the relationship between symptoms of depression and high-sensitivity crp.
\newblock \emph{International journal of obesity}, 37\penalty0 (1):\penalty0 S38--S43, 2013.

\bibitem[Wang et~al.(2023)Wang, He, Gao, and Calmon]{wang2023aleatoric}
Hao Wang, Luxi He, Rui Gao, and Flavio Calmon.
\newblock Aleatoric and epistemic discrimination: Fundamental limits of fairness interventions.
\newblock In \emph{Thirty-seventh Conference on Neural Information Processing Systems}, 2023.

\bibitem[Wang et~al.(2021{\natexlab{a}})Wang, Liu, and Levy]{wang2021fair}
Jialu Wang, Yang Liu, and Caleb Levy.
\newblock Fair classification with group-dependent label noise.
\newblock In \emph{Proceedings of the 2021 ACM conference on fairness, accountability, and transparency}, pages 526--536, 2021{\natexlab{a}}.

\bibitem[Wang et~al.(2019)Wang, Zhang, Liu, Pan, Hu, and Zhu]{wang2019acoustic}
Jingying Wang, Lei Zhang, Tianli Liu, Wei Pan, Bin Hu, and Tingshao Zhu.
\newblock Acoustic differences between healthy and depressed people: a cross-situation study.
\newblock \emph{BMC psychiatry}, 19:\penalty0 1--12, 2019.

\bibitem[Wang et~al.(2007)Wang, Aguilar-Gaxiola, Alonso, Angermeyer, Borges, Bromet, Bruffaerts, De~Girolamo, De~Graaf, Gureje, et~al.]{wang2007use}
Philip~S Wang, Sergio Aguilar-Gaxiola, Jordi Alonso, Matthias~C Angermeyer, Guilherme Borges, Evelyn~J Bromet, Ronny Bruffaerts, Giovanni De~Girolamo, Ron De~Graaf, Oye Gureje, et~al.
\newblock Use of mental health services for anxiety, mood, and substance disorders in 17 countries in the who world mental health surveys.
\newblock \emph{The Lancet}, 370\penalty0 (9590):\penalty0 841--850, 2007.

\bibitem[Wang et~al.(2022)Wang, Wang, Li, Zhang, and Wang]{wang2022online}
Yiding Wang, Zhenyi Wang, Chenghao Li, Yilin Zhang, and Haizhou Wang.
\newblock Online social network individual depression detection using a multitask heterogenous modality fusion approach.
\newblock \emph{Information Sciences}, 609, 2022.

\bibitem[Wang et~al.(2021{\natexlab{b}})Wang, Wang, Beutel, Prost, Chen, and Chi]{wang2021understanding}
Yuyan Wang, Xuezhi Wang, Alex Beutel, Flavien Prost, Jilin Chen, and Ed~H Chi.
\newblock Understanding and improving fairness-accuracy trade-offs in multi-task learning.
\newblock In \emph{Proceedings of the 27th ACM SIGKDD Conference on Knowledge Discovery \& Data Mining}, pages 1748--1757, 2021{\natexlab{b}}.

\bibitem[Wei et~al.(2022)Wei, Peng, Roitberg, Yang, Zhang, and Stiefelhagen]{wei2022multi}
Ping-Cheng Wei, Kunyu Peng, Alina Roitberg, Kailun Yang, Jiaming Zhang, and Rainer Stiefelhagen.
\newblock Multi-modal depression estimation based on sub-attentional fusion.
\newblock In \emph{European Conference on Computer Vision}, pages 623--639. Springer, 2022.

\bibitem[Wick et~al.(2019)Wick, Tristan, et~al.]{wick2019unlocking}
Michael Wick, Jean-Baptiste Tristan, et~al.
\newblock Unlocking fairness: a trade-off revisited.
\newblock \emph{Advances in neural information processing systems}, 32, 2019.

\bibitem[Wiens et~al.(2019)Wiens, Saria, Sendak, Ghassemi, Liu, Doshi-Velez, Jung, Heller, Kale, Saeed, et~al.]{wiens2019no}
Jenna Wiens, Suchi Saria, Mark Sendak, Marzyeh Ghassemi, Vincent~X Liu, Finale Doshi-Velez, Kenneth Jung, Katherine Heller, David Kale, Mohammed Saeed, et~al.
\newblock Do no harm: a roadmap for responsible machine learning for health care.
\newblock \emph{Nature medicine}, 25\penalty0 (9):\penalty0 1337--1340, 2019.

\bibitem[Williamson et~al.(2016)Williamson, Godoy, Cha, Schwarzentruber, Khorrami, Gwon, Kung, Dagli, and Quatieri]{williamson2016detecting}
James~R Williamson, Elizabeth Godoy, Miriam Cha, Adrianne Schwarzentruber, Pooya Khorrami, Youngjune Gwon, Hsiang-Tsung Kung, Charlie Dagli, and Thomas~F Quatieri.
\newblock Detecting depression using vocal, facial and semantic communication cues.
\newblock In \emph{Proceedings of the 6th International Workshop on Audio/Visual Emotion Challenge}, pages 11--18, 2016.

\bibitem[Xu et~al.(2020)Xu, White, Kalkan, and Gunes]{xu2020investigating}
Tian Xu, Jennifer White, Sinan Kalkan, and Hatice Gunes.
\newblock Investigating bias and fairness in facial expression recognition.
\newblock In \emph{Computer Vision--ECCV 2020 Workshops: Glasgow, UK, August 23--28, 2020, Proceedings, Part VI 16}, pages 506--523. Springer, 2020.

\bibitem[Yuan et~al.(2024)Yuan, Shi, Xu, Yang, Geng, and Rui]{yuan2024learning}
Hua Yuan, Yu~Shi, Ning Xu, Xu~Yang, Xin Geng, and Yong Rui.
\newblock Learning from biased soft labels.
\newblock \emph{Advances in Neural Information Processing Systems}, 36, 2024.

\bibitem[Zanna et~al.(2022)Zanna, Sridhar, Yu, and Sano]{zanna2022bias}
Khadija Zanna, Kusha Sridhar, Han Yu, and Akane Sano.
\newblock Bias reducing multitask learning on mental health prediction.
\newblock In \emph{ACII}, pages 1--8. IEEE, 2022.

\bibitem[Zhang and Yang(2021)]{zhang2021_multitask_survey}
Yu~Zhang and Qiang Yang.
\newblock A survey on multi-task learning.
\newblock \emph{IEEE Transactions on Knowledge and Data Engineering}, 34\penalty0 (12):\penalty0 5586--5609, 2021.

\bibitem[Zhang et~al.(2020)Zhang, Lin, Liu, and Mahmoud]{zhang2020multimodal}
Ziheng Zhang, Weizhe Lin, Mingyu Liu, and Marwa Mahmoud.
\newblock Multimodal deep learning framework for mental disorder recognition.
\newblock In \emph{15th IEEE International Conference on Automatic Face and Gesture Recognition (FG 2020)}, pages 344--350. IEEE, 2020.

\bibitem[Zheng et~al.(2023)Zheng, Yan, and Wang]{zheng2023two}
Wenbo Zheng, Lan Yan, and Fei-Yue Wang.
\newblock Two birds with one stone: Knowledge-embedded temporal convolutional transformer for depression detection and emotion recognition.
\newblock \emph{IEEE Transactions on Affective Computing}, 2023.

\end{thebibliography}
